\documentclass[10pt,letterpaper]{article}
\usepackage[top=0.85in,left=2.75in,footskip=0.75in]{geometry}

\usepackage{amsmath,amssymb}

\usepackage{changepage}

\usepackage{textcomp,marvosym}

\usepackage{nameref,hyperref}

\usepackage[right]{lineno}

\usepackage[numbers, square, sort&compress]{natbib}

\usepackage[nopatch=eqnum]{microtype}
\DisableLigatures[f]{encoding = *, family = * }

\usepackage[table]{xcolor}

\usepackage{array}

\newcolumntype{+}{!{\vrule width 2pt}}

\newlength\savedwidth

\raggedright
\usepackage[aboveskip=1pt,labelfont=bf,labelsep=period,justification=raggedright,singlelinecheck=off]{caption}

\usepackage[normalem]{ulem}

\makeatletter
\renewcommand{\@biblabel}[1]{\quad#1.}
\makeatother

\usepackage{lastpage,fancyhdr,graphicx}
\usepackage{epstopdf}

\usepackage{acronym}
\usepackage{cleveref}
\usepackage{hyperref}
\usepackage{float}
\usepackage{svg}
\usepackage{tcolorbox}
\newcolumntype{H}{>{\setbox0=\hbox\bgroup}c<{\egroup}@{}}

\fancyheadoffset[L]{2.25in}
\fancyfootoffset[L]{2.25in}
\begin{document}
\vspace*{0.2in}

\begin{flushleft}
{\Large
\textbf{Learning graph topology from metapopulation epidemic encoder-decoder} 
}
\newline
\\
Xin Li\textsuperscript{1},
Jonathan Cohen\textsuperscript{1},
Shai Pilosof\textsuperscript{2,3},
Rami Puzis*\textsuperscript{1}
\\
\bigskip
\textbf{1} Faculty of Computer and Information Science, Ben-Gurion University of the Negev, Be’er Sheva, South District, Israel
\\
\textbf{2} Department of Life Sciences, Ben-Gurion University of the Negev, Be’er Sheva, South District, Israel
\\
\textbf{3} The Goldman Sonnenfeldt School of Sustainability and Climate Change, Ben-Gurion University of the Negev, Be’er Sheva, South District, Israel
\\
\bigskip

%
%




* puzis@bgu.ac.il

\end{flushleft}


%
\section*{Abstract}
Metapopulation epidemic models are a valuable tool for studying large-scale outbreaks. 
With the limited availability of epidemic tracing data, it is challenging to infer the essential constituents of these models, namely, the epidemic parameters and the relevant mobility network between subpopulations. 
Either one of these constituents can be estimated while assuming the other; however, the problem of their joint inference has not yet been solved.
Here, we propose two encoder-decoder \acl{dl} architectures that infer metapopulation mobility graphs from time-series data, with and without the assumption of epidemic model parameters. 
Evaluation across diverse random and empirical mobility networks shows that the proposed approach outperforms the state-of-the-art topology inference. 
Further, we show that topology inference improves dramatically with data on additional pathogens. 
Our study establishes a robust framework for simultaneously inferring epidemic parameters and topology, addressing a persistent gap in modeling disease propagation.

%
\section*{Author summary}
Understanding how diseases spread in metapopulations is crucial for managing epidemics. In this study, we developed a deep learning approach that can uncover the hidden connections between subpopulations based solely on epidemic time-series data. Instead of assuming how people move between locations, our model learns these patterns directly from the infection records of multiple pathogens. We built an encoder-decoder framework that simultaneously learns both the structure of the mobility network and the parameters of the epidemic model. 
By testing our model on various simulated and real-world networks, such as transportation and regional mobility graphs, we found that it can accurately reconstruct the underlying connections and outperform existing inference methods. Our findings show that analyzing data from multiple disease pathogens greatly improves inference accuracy. 
This work presents a novel approach to elucidating population movement and disease transmission patterns, which may aid researchers and public health officials in better understanding and controlling future epidemics.

\linenumbers
\nolinenumbers
\section*{Introduction}

The outbreaks of infectious diseases such as SARS \cite{mclean2005sars}, influenza A (H1N1) \cite{fraser2009pandemic}, Ebola \cite{chowell2014transmission}, MERS \cite{cowling2015preliminary}, and COVID-19 \cite{wolfel2020virological} pose significant public health and the economic challenges \cite{kaye2021economic}. 
The growth of cities and well-connected transportation networks contributes to the rapid spread of pathogens\footnote{Throughout this paper, the term pathogen refers to a parasitic microorganism (e.g., bacteria, viruses, fungi) that causes disease in its host.} \cite{brockmann2006scaling}.
Therefore, it is crucial to understand how the mobility of the population influences the transmission of diseases at large scales in order to devise effective containment and intervention policies.

The theory of complex networks offers a useful framework for studying this problem \cite{estrada2012structure, wang2003complex}.
Specifically, \textit{metapopulation networks} define subpopulations (e.g., cities, countries) as nodes and the mobility between them as links \cite{wang2014spatial,meyers2005network}.
While networks provide the backbone structure on which a pathogen spreads, epidemic models describe the underlying transmission mechanisms. 
Hybrid metapopulation epidemic models \cite{murphy2021deep, wang2014spatial} provide a good trade-off between the scalability of the compartmental models \cite{brauer2008compartmental,ozmen2016analyzing}, which assume fully mixed populations, and the flexibility of agent-based models \cite{bisset2009epifast, venkatramanan2018using, zhao2015simnest, keeling2008modeling}, which focus on individuals and their contacts. 
Hybrid models provide valuable insights into the propagation of infectious diseases between subpopulations at large scales \cite{colizza2007reaction}.
For example, Brockmann et al.~\cite{brockmann2013hidden} and Hufnagel et al.~\cite{hufnagel2004forecast} used a global aviation graph to analyze the propagation of SARS and H1N1, respectively,  in the global city metapopulation, Wesolowski et al. \cite{wesolowski2012quantifying} used a cellphone user mobility graph to analyze malaria propagation in an inter-settlement metapopulation of Kenya, demonstrating successful applicability and a high degree of predictability. 
Here, we focus on metapopulation networks composed of spatially separated fully mixed populations, within the epidemic \acf{sir} model.

Predicting the spread of epidemics in real-world metapopulations poses significant challenges, primarily due to unknown disease transmission networks  and epidemic model parameters.
The topology inference problem \cite{wang2018inferring, barabasi1999emergence,  matsuki2025network} focuses on uncovering the disease transmission network, assuming known epidemic model parameters. 
The epidemic model estimation problem \cite{o1999bayesian, taghizadeh2022seir, calciano2025epidemic} aims to determine the unknown parameters of the epidemic model, assuming that the disease transmission network aligns with a known mobility graph.  
Solving both problems is essential for accurately modeling real-world epidemic spread. 


\paragraph{Uncovering the disease transmission network.}
One approach to account for the unknown structure of a metapopulation network is to use proxies.
For instance, cell phone data provides the time and location of people in the same region, providing a proxy for human proximity. Indeed, cell phone user mobility networks can explain the spread of malaria \cite{wesolowski2012quantifying}.
However, using proxy networks is challenging because the network used should match the transmission mode of the pathogen.
For example, the airline travel network was used to estimate how travel restrictions would affect the spread of H1N1 influenza \cite{salathe2010high, bajardi2011human}, but it cannot be used to study the dynamics of HIV \cite{keeling2008modeling,silk2018contact, magiorkinis2016global}.

Rather than relying on proxy networks, it is possible to infer the metapopulation network directly from epidemic data.
However, the inference process is challenging due to the large number of unobserved variables (unknown links).
To improve the accuracy of network topology inference, researchers often introduce assumptions based on partial knowledge of the network structure.
For example, Wang et al.\cite{wang2018inferring} assumed a power-law distribution in the mobility graph, while others assumed a gravity model to assign link weights \cite{deng2020cola}.
Geographic bordering relationships were also employed to infer the transmission probability \cite{yu2023spatio}.
However, these assumptions can oversimplify the complexities associated with modern transportation systems, diverse travel methods, and dynamic population flows.

\paragraph{Estimating the epidemic model parameters. }
In the literature, researchers have utilized both probabilistic and deterministic frameworks to infer disease dynamics from aggregate data. O'Neill et al.
\cite{o1999bayesian}, and Taghizadeh et al.\cite{taghizadeh2022seir} employ Bayesian inference and nonlinear least-squares methods, respectively, to estimate epidemic parameters.
These methodologies treat the subpopulation of interest as a fully mixed unit, assuming that individuals interact homogeneously or function independently, without requiring detailed contact network data.
Extending this to networked systems, Calciano et al. \cite{calciano2025epidemic} use Markov Chain Monte Carlo methods to fit epidemic models to metapopulation networks and evaluate how specific topologies affect the accuracy of epidemic parameter estimates.


\begin{tcolorbox}[
       colback=gray!5,
       colframe=gray!40!black,
       width=1.05\textwidth,
       arc=1mm,
       boxrule=0.5pt,
       center,
       halign=center,
       left=1pt, top=1pt, right=1pt, bottom=1pt,
   ]
To the best of our knowledge, this is the first attempt to infer metapopulation topologies without making assumptions about their structure and epidemic parameters.
\end{tcolorbox}

In this paper, we advance the topology inference state of the art by 
introducing a topology inference encoder-decoder deep learning framework that: (1) requires no prior assumptions about network structure, (2) jointly infers both topology and \ac{sir} epidemic parameters, and (3) leverages multi-pathogen data to improve identifiability. 
Our \ac{dtef} architecture combines a \ac{dti} encoder with a novel parameter-estimation approach via the \ac{efb} decoder.
We evaluate our method on both synthetic and real-world graphs. 
The main contributions of this paper are summarized as follows:
\begin{enumerate}
    \item We propose (\ac{dtef}), the first framework capable of jointly inferring both the network topology and the epidemic parameters from metapopulation epidemic time-series data. 
    \item We demonstrate that topology inference accuracy improves with the number of independent pathogens. 
    \item Provided the epidemic parameters, \ac{dtef} exhibits state-of-the-art performance in topology inference. 
\end{enumerate}

The paper is organized as follows: in Section II we introduce the SIR model, and in Section III we discuss the relevant learning tasks within the multi-pathogen metapopulation SIR model.
We describe our \ac{dtef} model for epidemic topology inference in Section IV.
In Section V we perform numerical simulations and present the experimental results.
Finally, in Section VI we summarize our findings and present our plans for future work. 
In Appendix~\nameref{subsec:theoretical_proofs} we provide theoretical proofs, in Appendix~\nameref{subsec:realgraphattribs} we provide details on the real-world graphs used in the paper, and in Appendix~\nameref{subsec:supp exp results} we provide supplementary experimental results.

\section*{Preliminaries and Problem Formulation}

\subsection*{Metapopulation \acs{sir} model}

\begin{figure*}[t]
    \includegraphics[width=1\textwidth]{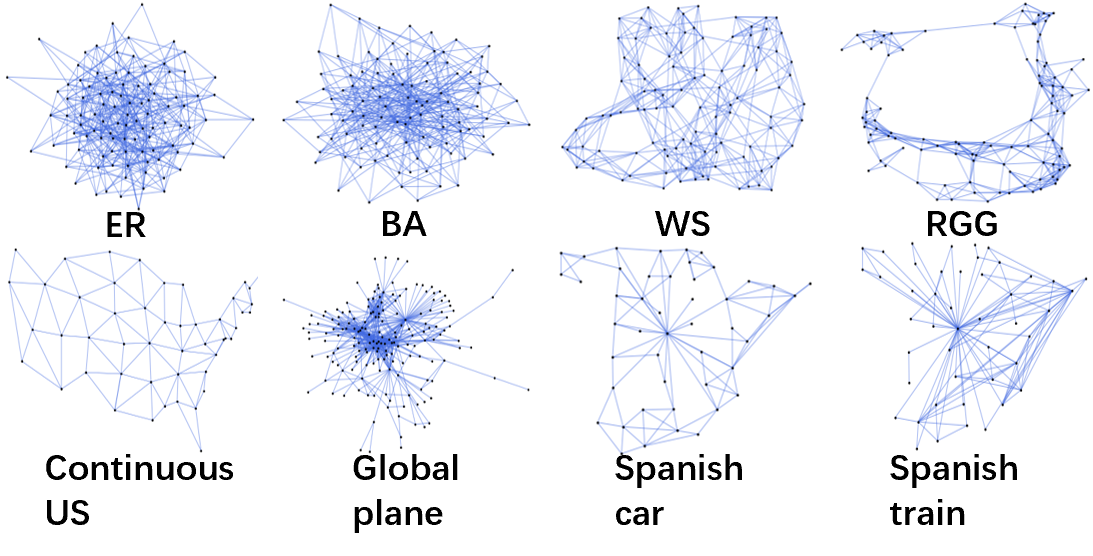}
    \caption{
    \textbf{Representative graph topologies used in this study.}
    Top row: synthetic random graphs - \acf{er}, \acf{ba}, \acf{ws}, \acf{rgg}.
    }
    \label{fig:topology}
\end{figure*}

The metapopulation \acs{sir} model \cite{murphy2021deep, wang2014spatial} considers $n$ populations that are connected by the migration of individuals between them. 
Each subpopulation has its own set of \acs{sir} compartments $X_i=(S_i, I_i, R_i)|i \in [1, n]$ denoting the proportion of susceptible, infectious, and recovered individuals in subpopulation $i$, and the following sum holds: $S_i+I_i+R_i=1$.
These subpopulations are interconnected through the migration of individuals, resulting in the metapopulation flow graph $G=(V, E)$, 
where $V$ and $E$ are the sets of nodes (subpopulations) and links that connect them, with sizes $|E|=m$ and $|V|=n$. 
Representative examples of these graphs are illustrated in \cref{fig:topology}, while full details are provided in the Experimental Setup.

\begin{figure*}[t]
\minipage{0.5\textwidth}
  \includegraphics[width=\linewidth]{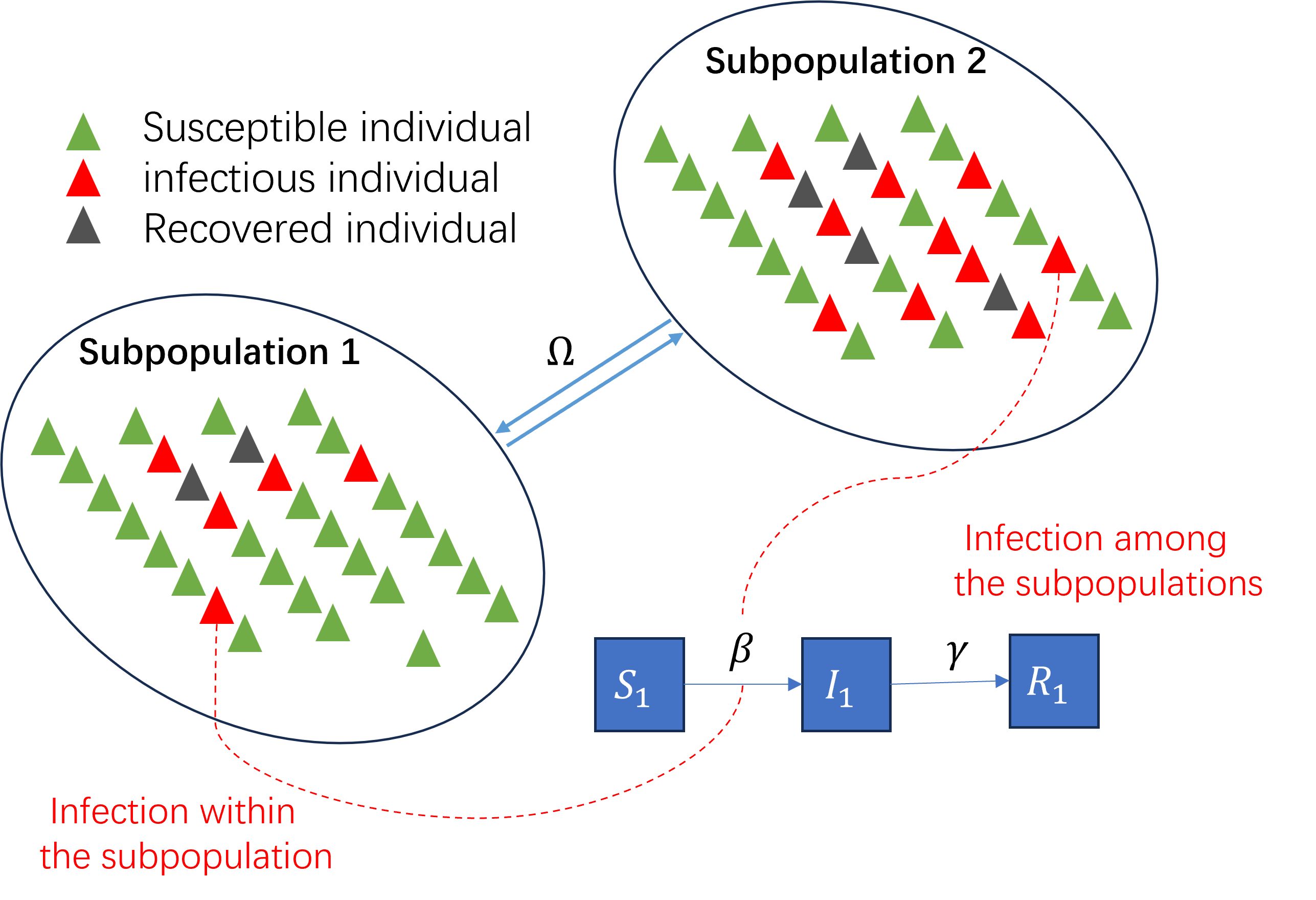}
\endminipage\hfill
\minipage{0.5\textwidth}
  \includegraphics[width=\linewidth]{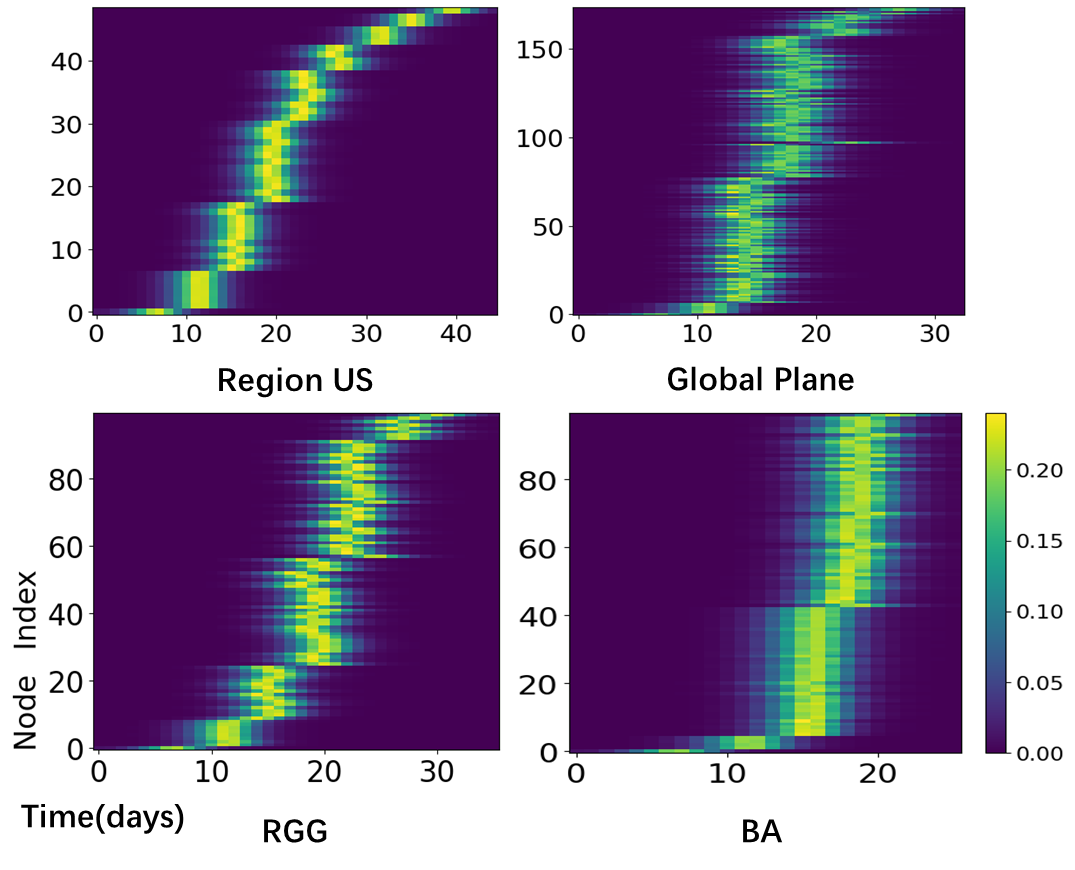}
\endminipage\hfill
\caption{ Left: A metapopulation epidemic process showing the infection both within and among the populations. The blue blocks indicate the state evolution of subpopulation 1. Right: An example of one pathogen metapopulation epidemic time-series data $\Delta \bar S$ in typical \acs{ba} and  \acs{rgg} graphs with 100 nodes, as well as empirical contiguous US \cite{nr}, where the nodes are geometric states (two nodes establish a link if they share a border), and global air transportation graphs, which represent the commute between two airports. 
    The x-axis represents the number of days since the start of the epidemic.
    The nodes are ordered along the y-axis according to the distance from the seed node. 
    The color indicates the fraction of daily new cases $\Delta \bar S$. }
\label{fig: crossNodes_dataExample}
\end{figure*}

Within each population, individuals are assumed to mix homogeneously.
In Figure~\ref{fig: crossNodes_dataExample} (left), red infected individuals can infect others in their population and migrate between subpopulations, introducing the potential for the pathogen spread across the metapopulation.
The rate of migration $MobilityRate$ between any two subpopulations is represented with a mobility adjacency matrix $A$ whose element $A_{ij}$ denote the fraction of individuals from subpopulation $V_j$ traveling to subpopulation $V_i$.
For node $V_i$, the population size $P_i$ is assumed to remain constant.
We assume that individuals arriving at a new subpopulation have an equal chance of interacting with any individual in the destination subpopulation. 
Let $\Omega_{ij}=P_j \times A_{ij}$ denote the number of people traveling from node $v_j$ to node $v_i$.
In Figure~\ref{fig: crossNodes_dataExample} (left), all individuals in subpopulation 1 can commute to subpopulation 2 with the same chance and vice versa. 
The state can only evolve from S to I to R.
There are two constant parameters which are used to describe an \acs{sir} epidemic pathogen: transmission rate $\beta$ and recovery rate $\gamma$.
Using the mobility adjacency matrix $A$, the vector of population sizes $P$, and \acs{sir} epidemic parameters $\beta, \gamma$, we can define an SIR metapopulation-level model.
These parameters are also assumed constant during the time span $T$ for one pathogen. 

Given the assumptions described above, $S$,$I$,$R$ have the following dynamics \footnote{In what follows, all vectors are assumed to be row vectors unless specified otherwise.}:

\begin{equation}
\label{eq:diffusion}
\left(\begin{array}{l}
 S(t+1)
\\ I(t+1)
\\ R(t+1)\end{array}\right)=
\left(\begin{array}{l}
{S(t)-S(t)\odot\tilde{\alpha}(t)} 
\\ {{I( t)-I(t)\gamma+S(t)\odot\tilde{\alpha}(t)}}
\\ {{R( t)+I(t)\gamma}}\end{array}\right)
\end{equation}
where $t\in[0,T-1]$ is the time in days and $T$ is the entire time span of epidemic in our study.$\tilde{\alpha}(t)$ is the cumulative infection rate, given by
\begin{equation}
\label{eq:diffusion_matrix}
\tilde{\alpha}(t)=\alpha(I(t), P)\cdot (\boldsymbol{I}+\frac{d^T\cdot \Omega}{(\sum_{dim=1}\Omega)^T}) 
\end{equation}
where $\boldsymbol{I}$ is an identity matrix and $d=\sum_{dim=1} A$ is the in-degree of adjacency $A$. The function $\alpha(I(t), P)$ in Eq.~\eqref{eq:diffusion_matrix} corresponds to the infection rate per day, at which an individual is infected by a visitor from a neighboring subpopulation with $I(t)\odot P$ infected people in it, and is equal to:

\begin{equation}
\label{eq:alpha}
\alpha(I(t), P)=1-\left(1-{\frac{\beta}{P}}\right)^{I(t)\odot P}
\end{equation}
Assuming that population $P$ is numerically much larger than $ \beta$ and $I$, we obtain the following computationally-convenient approximate form (see Appendix~\nameref{subsec:alpha approx}):
\begin{equation}
\label{eq:alpha-app}
\alpha(I(t), P)\approx1-e^{-{\beta}I(t)}
\end{equation}

Finally, it is worth noting that the term $\alpha(I(t), P)\cdot \boldsymbol{I}$ in Eq.~\eqref{eq:diffusion_matrix} is the infection rate within a subpopulation, whereas $\alpha(I(t), P)\cdot\frac{d^T\odot \Omega}{(\sum_{dim=1}\Omega)^T}$ is the infection rate among subpopulations.
\subsection*{Multi-pathogen metapopulation \acs{sir} model}
The multi-pathogen metapopulation \acs{sir} model in this paper extends the metapopulation model in Eq.~\eqref{eq:diffusion} as a superposition of $k$ pathogens with the same transmission model, whose dynamics can be described using \acs{sir} dynamics.
We assume that these pathogens are spread across the same metapopulation $G$, whose mobility adjacency matrix is $A$, but differ in their \acs{sir} parameters and seed node \footnote{For clarity, the term 'seed node' indicates the node where the first documented patient appears during an epidemic.} \cite{kretzschmar1996measures, li2021modeling}.
Because our main objective is inferring the metapopulation structure, we disregard cross-immunity effects between pathogens (see, e.g., \cite{kamo2002effect, adams2007influence}). Hence, their dynamics are independent. Therefore,  each pathogen has its set of \acs{sir} compartments in each population $X_i(l)=(S_i(l), I_i(l), R_i(l))$. 

$X_i(l)$ denotes the \acs{sir} compartments of subpopulation $i$ for pathogen $l$. At the  metapopulation level, Eqs.~\eqref{eq:diffusion},~\eqref{eq:diffusion_matrix}, and~\eqref{eq:alpha-app} can be directly applied.

In the multi-pathogen metapopulation \acs{sir} model, $\beta$ and $\gamma$ are $k\times 1$ vectors $\hat{\beta},\hat{\gamma}$ denoting the epidemic parameters of different pathogens, where $\hat{\beta}(l)$ is the $\beta$ parameter of the $l$-th pathogen; the same notation applies for $\hat{\gamma}$.

\subsection*{Infection matrix}
In accordance with Eq.~\eqref{eq:diffusion_matrix}, we define the infection matrix Z as:

\begin{equation}
\label{eq:infection_matrix}
\begin{array}{l} Z=\boldsymbol{I}+\frac{d^T\odot \Omega}{(\sum_{dim=1}\Omega)^T}
\\=\boldsymbol{I}+\frac{(\sum_{dim=1}A)^T\odot A\odot P}{(\sum_{dim=1}(A\odot P))^T}
\end{array}
\end{equation}
where $Z_{ij}$ represents the extent to which the infection rate of node $i$ affects node $j$ in an epidemic, hence the name infection matrix.
Since the mobility matrix $A$ and population sizes $P$ are fixed, $Z$ is also fixed during the epidemic.
In addition, similar to the mobility adjacency matrix $A$, the infection matrix $Z$ is also sparse (see Appendix~\nameref{subsec:sparsity}).

Eq.~\eqref{eq:infection_matrix} is reversible (see Appendix~\nameref{subsec:reversibility}):
\begin{equation}
\label{eq:infection_reverse}
A=\frac{(Z-\boldsymbol{I}) \odot(\sum_{dim=1}(Z-\boldsymbol{I}))^T}{P\odot (\sum_{dim=1}\frac{Z-\boldsymbol{I}}{P})^T} 
\end{equation}
That is, given infection matrix $Z$ and population sizes $P$, we can compute matrix $A$ using Eq.~\eqref{eq:infection_reverse}.

\subsection*{Formal problem definition}
\begin{figure*}[t!]
    \includegraphics[width=1\textwidth]{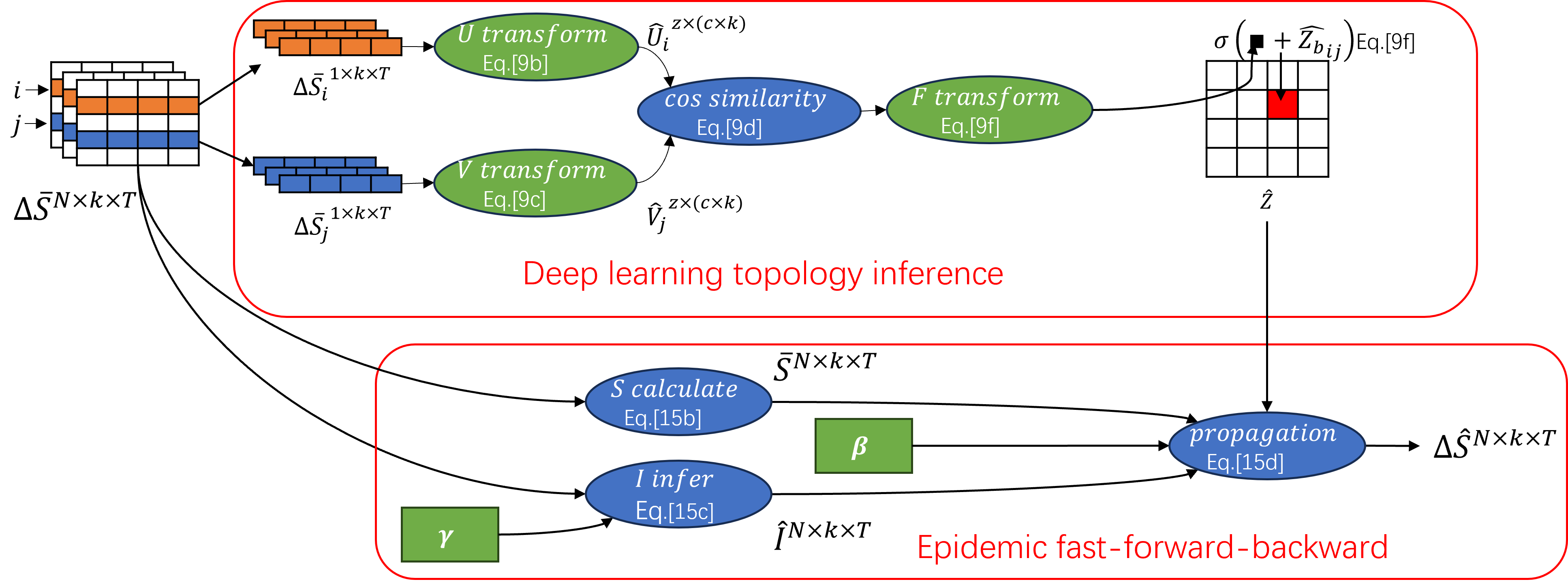}
    \caption{
    \textbf{Architecture of the self-supervised DTEF model.}
    The input is the daily new infection data $\Delta\bar{S}$. 
    The encoder extracts node--pathogen embeddings and infers the infection matrix $\hat{Z}$, 
    and the decoder uses $\hat{Z}$ to reconstruct the predicted daily new infections $\Delta\hat{{S}}$.
    Green ovals and rectangles denote modules with learnable parameters, blue ovals denote parameter-free operations, 
    the black box in $\sigma$ is the scalar output of $F$ transform operation, 
    $\sigma$ is the sigmoid activation enforcing $\hat{Z}_{ij}\in[0,1]$(red cell),
    arrows indicate input sequences, and the highlighted red cell shows a specific reconstructed entry of $\Delta\hat{S}$.}
    \label{fig:full_process}
\end{figure*}

Learning graph topology from multi-pathogen metapopulation epidemics requires fitting the metapopulation graph topology, using the epidemic time-series data collected from the epidemic during a specified time period.
Typically, epidemic data includes the number of daily new cases of infectious individuals in each subpopulation.
In our work, we use Eq.~\eqref{eq:diffusion} to simulate and collect the daily new infectious cases denoted as $\Delta \bar S^{n\times k\times T}$. 
We use $\Delta \bar S_i(l,t)$ to refer to the fraction of the population of node $i$ infected by pathogen $l$ at time $t$. 
Figure~\ref{fig: crossNodes_dataExample} (right) presents sample metapopulation time-series data of one pathogen of daily new cases in \acs{ba} and \acs{rgg} graphs. We also use two empirical networks (the contiguous US and global plane).
In those plots, the node index (y-axis) is ordered according to the shortest path distance from the seed node, the x-axis represents the number of days since the seed node occurred, and the color indicates the epidemic behavior (daily new cases). 
In particular, we see that in the vicinity of the seed node, the BA graph contains $\sim$3-4 distinctive subpopulations based on the number of neighbors of initial nodes, forming clusters that have different distances from the seed node
and whose nodes exhibit similar epidemic behavior. In contrast, the epidemic dynamics applied to an \acs{rgg} graph provide a richer structure for topology reconstruction (see, e.g., the real graphs contiguous US and Global Plane for comparison). 
In the rest of the paper, we will show that given the \acs{sir} dynamics (whose epidemic parameters are learned from the epidemic diffusion data), it is possible to reconstruct the structure of the underlying disease-propagation network.


The number of daily new cases, as deduced from Eq.~\eqref{eq:diffusion}, are:
\begin{equation}
\label{eq: propagation}
\begin{array}{l} -\Delta \bar{S}(t)=S(t)-S(t+1)
\\ \approx S(t) \odot ( 1-e^{-{\hat{\beta}} \cdot I(t)} )\cdot Z
\end{array}
\end{equation}

In Eq.~\eqref{eq: propagation}, given the daily new case data $\Delta \bar{S}$, the task of our proposed method is to infer the infection matrix $Z$ as well as the corresponding epidemic parameters. 
If $-\Delta \bar{S}(t)$ were known and $ S(t) \odot ( 1-e^{-{\hat \beta  } \cdot I(t)} )$ were treated as a fixed transformation, this is a typical linear regression problem, which has a numerical solution. 
However, $ S(t) \odot ( 1-e^{-{\hat \beta } \cdot I(t)} )$ contains unknown parameters, giving rise to nonlinear effects which we will address using \ac{dl} techniques.
Once the infection matrix $Z$ is inferred, we can calculate the mobility matrix $A$ via Eq.~\eqref{eq:infection_reverse}. 
Two of our method's sub-tasks are infection matrix $Z$ learning and epidemic parameters learning.  

We propose \ac{dti} and \ac{efb} to perform these two tasks, evaluating the performance of our refined models in section~\nameref{sec:experiments}.

\section*{Methods}
Figure~\ref{fig:full_process} presents the architecture of the \ac{dtef} model, where the green blocks contain learnable parameters. 
The input of this \ac{dl} model is the ground-truth daily new cases $\Delta \bar{S}$, and the output is the predicted daily new cases $\Delta \hat{S}^{n\times k\times T }$.
Our \ac{dl} model comprises two sub-models: \ac{dti} and \ac{efb}.
The \ac{dti} model infers the infection matrix by learning the potential connections between two populations, while the \ac{efb} model provides an efficient means of back-propagating in the diffusion process, leveraging the computed infection matrix to predict the daily new cases $\Delta \hat{S}$. 

\subsection*{\ac{dti} model}

In this subsection, we describe our \ac{dti} model for inferring the topology matrix, which is a more effective alternative to the straightforward \ac{fti} technique.
In the latter method, one considers $P(\Delta \hat{S}| \hat{Z})$ as the probability of observing $\Delta \hat{S}$ given $\hat{Z}$, such that topology inference involves maximizing the expectation to observe $\Delta \hat{S}$ given $\hat{Z}$: i.e., $\max \quad E_{\hat{Z}}[log P(\Delta \bar{S}=\Delta \hat{S}|\hat{Z})]$.
This can be achieved by computing the prior observation $\Delta \hat{S}$ and back-propagating the gradient to update the posterior topology $\hat{Z}$.
Thus, the \Ac{fti} model is shown in Eq.~\eqref{eq: fast topology inference}, where infection matrix $W_z$ is an $n\times n$ learnable weight, and infection matrix $\hat{Z}$ can be directly learned through the gradient back-propagation:
\begin{align}
\label{eq: fast topology inference}
\notag \textbf{input}&: None 
\\
\hat{Z}&=\sigma(W_z)
\\
\notag
\textbf{output}&: \hat{Z}
\end{align}


The \Ac{fti} model directly parameterizes the infection matrix 
$\hat{Z} \in [0,1]^{n \times n}$ and learns its entries independently through back propagation. 
However, this parameterization provides no mechanism to capture shared patterns among nodes, 
and the search space $[0,1]^{n \times n}$ can be high-dimensional for large $n$. 
Motivated by these observations, we introduce the \ac{dti} model, 
which maps the temporal infection sequences into latent embeddings and computes pairwise 
similarities to infer $\hat{Z}$. This shared embedding couples the entries of $\hat{Z}$ 
and reduces the effective search space.

Neural networks, particularly \ac{dl} models, are powerful tools for learning complex and hierarchical embeddings from data.
In what follows, we design a \ac{dl} model in the spirit of variational inference to infer the infection matrix. 
\footnote{In the variational method, one seeks an easy-to-solve variational distribution $q_\phi(\hat{Z}|\Delta \hat{S})$ to approximate $\hat{Z}$ given $\Delta \hat{S}$, where $\phi$ are the parameters of this distribution, such that the objective function is given by  $\max \quad E_{q_\phi(\hat{Z}|\Delta \hat{S})}[log P(\Delta \bar{S}=\Delta \hat{S}|\hat{Z})]$. In our proposed \acf{dti}, a technique which is also called learning to optimize, we use neural networks as parameters $\phi$ accordingly, computing the prior observation $\Delta \hat{S}$ and back-propagating the gradient to update the parameters of the variational distribution $q_\phi$.}

Epidemic diffusion progress in a graph is traceable, meaning that any newly infected node must have at least one infected neighbor from which the contagion originated.
Hence, the epidemic dynamics of connected nodes are correlated.
For example, assume that $A_{1,0}$ is larger than $A_{2,0}$ and that the seed node initially occurs on node $0$. In that case, at the early stage of the epidemic, the number of infected individuals in $v_1$ will be more similar to that of $v_0$ than the number of $v_2$.

Based on this observation, we design an \ac{dl} model to infer $Z$.
The \ac{dti} model reads:

\begin{subequations}
\label{eq: inference_model_all}
\begin{align}
\textbf{input}&: \Delta {\bar S}|{i,j\in[1, n], l\in[1, k]}
\\
\label{eq: inference_model}
 \hat U_i(l)&=W_u\cdot (\Delta \bar S_i(l))+b_u
\\ 
\label{eq: inference_model1}
\hat V_j(l)&=W_v\cdot(\Delta \bar S_j(l))+b_v
\\ 
\label{eq: inference_model2}
Sim_{ij}(l)&=\frac{\sum \hat{U}_i(l)\odot \hat{V}_j(l)}{|\hat{U}_i(l)|\times |\hat{V}_j(l)|}
\\ 
\label{eq: inference_model21}
E\{Sim_{ij}\}&= \frac{\sum_l(Sim_{ij}(l))}{k}
\\
\label{eq: inference_model3}
\hat{Z}_{ij}&=\begin{cases}
\sigma(W_F\cdot E\{Sim_{ij}\}+\hat{Z_b}_{ij}) & {i\neq j}\\
    1 & {i= j}
  \end{cases}
\\
\label{eq: inference_model31}
\hat{Z}&= \sigma(\zeta)\hat{Z}+(1-\sigma(\zeta))\hat{Z}.T
\\
\textbf{output}&:\hat{Z}
\end{align}
\end{subequations}

In Eq.~\eqref{eq: inference_model}, the learnable weight $W_u^{(z\times c) \times T}$ transforms the $a-$th pathogen epidemic data of the $i$-th node $\Delta \bar S(a, i)^{T\times 1}$ into an embedding $U_i(l)^{(z\times c) \times k}$, where $z$ is a hyperparameter denoting the dimension of embedding features (embedding), which we set to $30$ \footnote{The embedding size was carefully chosen based on preliminary experiments.}. In addition, $c$ is the hyperparameter of feature channels which we set to 5.
Similarly, in Eq.~\eqref{eq: inference_model1}, the learnable weight $W_v^{z\times c \times T}$ transforms the $a-$th pathogen epidemic data of the $j$-th node $\Delta \bar S_j(l)^{T\times 1}$ into an embedding $V_j(l)^{z\times c \times k}$.
$b_u, b_v$ are scalar learnable biases.

Eq.~\eqref{eq: inference_model2} calculates the cosine similarity embedding $Sim_{ij}(l)^{c\times k}$ between $\hat U(a, i)$ and $\hat V(a, i)$ on their embedding dimension. 
Eq.~\eqref{eq: inference_model21} averages the cosine similarity embedding across all pathogens into $E\{Sim_{ij}\}^{c}$.
Using a learnable weight $W_F^{1\times c}$, Eq.~\eqref{eq: inference_model3} fuses the averaged similarity embedding into a scalar value.
Eq.~\eqref{eq: inference_model31} leverages the information that the incoming and outcome mobility rate are similar.
$\sigma$ is the sigmoid function.
Because the similarity cannot cover all potential features of epidemic diffusion in a graph, we add a learnable infection bias $\hat{Z_b}^{k\times k}$ to improve its state space.

In the above formulation, $W_u, b_u, W_v, W_F, b_F, \hat{Z_b}, \zeta$ are learnable parameters.
Thus, we are now in a position to generate the whole infection matrix $\hat{Z}^{k\times k}$ by computing all $i,j$ pairs. In section~\nameref{sec:experiments} we present our experimental results which demonstrate the effectiveness of this method.

To conclude, our approach essentially transforms the original ill-defined regression problem into a feature extraction and pairwise embedding problem.
The variable space for the regression is $n^2$, whereas for the feature extraction, it is $T\times z\times c$.
In fact, $T\times z\times c$ is typically much smaller than $n^2$, especially when $n>100$.
Thus, solving the feature extraction and pairwise embedding tasks is easier than solving the original regression problem.
Meanwhile, because our method has a relatively small variable space, it is less prone to overfitting and local optima \cite{hastie2009elements}.

\subsection*{\ac{efb} model}



In this subsection, we describe our \acf{efb} model for inferring the epidemic \acs{sir} parameters, which is a fast (i.e., computationally efficient) alternative to the traditional \ac{esc} technique.
In Eq.~\eqref{eq: propagation}, $S_0, I_0$ are pre-specified values, and $\hat{\beta}, \hat{\gamma}$ are learnable parameters.
\Ac{esc} sequentially computes the daily new infectious cases $\Delta \hat{S}(t)|{t\in [1, T]}$. 
Similar to RNN training, by comparing the ground truth $\Delta S$ to the predicted $\Delta \hat{S}$, it back-propagates the gradients to update the epidemic parameters $\hat{\beta}, \hat{\gamma}$. 
Most epidemic prediction methods predict the epidemic using similar principles \cite{wang2018inferring,mao2023mpstan,deng2020cola}. Therefore, the \ac{esc} model reads:

\begin{align}
\label{eq: epidemic sequential computation}
\notag
\textbf{input}&: \Delta {\bar S}(1)
\\
\notag
\hat{S}(t)&= \hat{S}(t-1)- \Delta \hat{S}(t)
\\
\hat{I}(t)&= \hat{I}(t-1)+ \Delta \hat{S}(t) - I(t)\hat{\gamma}
\\
\notag
\Delta \hat{S}(t)&=\hat{S}(t-1)\odot (1-e^{-{ \hat{\beta} } \cdot \hat I(t)})\cdot \hat{Z}
\\
\notag
\textbf{Output}&: \Delta \hat{S}(t)|{t\in[1, T]}
\end{align}
Unfortunately, the \ac{esc} model depends heavily on the accuracy of the initial values $S_0, I_0$. Furthermore, intermediate values $\hat{S}(t), \hat{I}(t)$ will tend to exacerbate the error which is computed based on the preceding uncertain time value $\hat{S}(t-1), \hat{I}(t-1)$, resulting in an undesirable gradient.
Additionally, this method suffers from a similar flaw as the gradient vanishing problem, hindering its convergence.
In practice, we found that the sequential computation is slow and does not provide accurate inference results.

 We, therefore, propose a more efficient computation method.
 By using Eq.~\eqref{eq: propagation}, given the ${S}, {I}$ vector pairs, we can easily compute the decoder output.
Ground truth $\bar {S}$ can also be accurately calculated by accumulating the daily new cases: $\bar {S}(t)=1-\sum_{i=0}^{t} \Delta \bar S(i)$. Rewriting this in matrix form, we have: 

$$\bar {S}= 1-L\cdot \Delta \bar S^T$$
where $L_{T\times T}$ is a lower triangular one matrix:
\begin{equation}
L= \begin{pmatrix}
1\ \ \  0 \ \ \  0\ \ \  0\ \ \ ... \\
1\ \ \  1\ \ \  0\ \ \  0\ \ \ ... \\
1\ \ \  1\ \ \  1\ \ \  0\ \ \ ... \\
1\ \ \  1\ \ \  1\ \ \  1\ \ \ ... \\
... \end{pmatrix}
\end{equation}
To calculate the vector ${\hat{I}}$, we observe that $\hat{I}(t)+\hat{R}(t)=1-\bar{S}(t)$. If we set $\hat{\gamma}$ to be learnable, from Eq.~\eqref{eq:diffusion} we can obtain the predicted $\hat{I}_t$:
\begin{equation}
\label{eq:diffusion3}
\hat{I}(t)=\sum_{t_i=1}^t{(1-\hat{\gamma})^{t-t_i}\Delta \bar{S}(t_i)}
\end{equation}
which can be written in a matrix form as:
\begin{equation}
\label{eq:diffusion4}
\hat{I}=e^{B\cdot ln(1-\hat{\gamma})}\cdot\Delta \bar{S}
\end{equation}
where $B_{T\times T}$ is a matrix similar to a triangular number sequence (here '$/$' means disregard value):
\begin{equation}
B= \begin{pmatrix}
0\ \ \  / \ \ \  /\ \ \  /\ \ \ ... \\
1\ \ \  0\ \ \  /\ \ \  /\ \ \ ... \\
2\ \ \  1\ \ \  0\ \ \  /\ \ \ ... \\
3\ \ \  2\ \ \  1\ \ \  0\ \ \ ... \\
... \end{pmatrix}
\end{equation}
In summary, we propose the following computationally efficient \ac{efb} model:
\begin{subequations}
\label{eq: process_all}
\begin{align}
\textbf{input}&: \Delta {\bar S}(t)|{t\in[1, T]}
\\
\label{eq: process}
\bar {S}&= 1- L\cdot \Delta \bar S^T
\\ 
\label{eq: process1}
\hat{I}&=e^{B\cdot ln(1-\hat{\gamma})}\cdot\Delta \bar{S}^T
\\ 
\label{eq: process2}
\Delta \hat{S}(t)&=\bar{S}(t-1)\odot (1-e^{-{\hat{\beta}} \cdot \hat I(t)})\cdot \hat{Z}
\\
\textbf{output}&: \Delta {\hat S}(t)
\end{align}
\end{subequations}
where Eq.~\eqref{eq: process} calculates ground-truth daily susceptible data $\bar{S}$, Eq.~\eqref{eq: process1} computes the predicted daily infectious data $\hat{I}$, and Eq.~\eqref{eq: process2} computes the predicted daily new cases $\Delta \hat{S}$.
Note that the infection matrix $\hat{Z}$ in Eq.~\eqref{eq: process2} emanates from the infection matrix inference module. 
In fact, the \ac{efb} model can perform fast parallel training.
Also, our model does not depend on the initial value $\Delta {\bar S}(1)$, which in turn further contributes to its robustness. 

This method of \ac{efb} can be applied to other population-level models composed of one diffusion process and multiple temporal transmission rates, e.g., \ac{seir} and \ac{sird}. 
By introducing temporal transmission rates as learnable parameters in addition to the constant population constraint, all states can be readily solved. 
Then we can insert all solved states into the diffusion process to compute the decoder output.
Finally, we update the diffusion parameters and temporal transmission rates through gradient descent.

\subsection*{Loss function}
By combining the infection matrix and an efficient computation module, given a ground truth $\Delta \bar{S}$, we can efficiently compute the predicted $\Delta \hat{S}$. In this subsection, we describe the loss function associated with the back-propagation.
The loss function is composed of two parts: prediction loss and minimal variance loss.
The prediction loss ensures an accurate prediction:
\begin{equation}
\label{eq:prediction_loss}
\|\Delta \hat{S}-\Delta \bar S\|_2
\end{equation}
where $\|\cdot\|_2$ is the L2 norm.

In training, to improve the search space, we extend the size of the epidemic parameters with the subpopulation dimension, such that $\hat{ \beta},\hat{\gamma}$ have dimension $k\times n$.
We randomly initialize \acs{sir} parameter vectors $\hat{\beta}, \hat{\gamma}$.  Since these  \acs{sir} parameter vectors should be the same, we subsequently minimize the vectors' variance:
\begin{equation}
\label{eq:variance_loss}
var(\hat{\gamma})+var(\hat{\mathcal \beta})
\end{equation}
where $var()$ is the mathematical variance.
Finally, the overall loss  can be computed as follows:
\begin{equation}
\label{eq: loss}
\mathcal{L}=\frac{\|\Delta \hat{S}-\Delta \bar S\|_2+var(\hat{\gamma})+var(\hat{\mathcal \beta})}{T}
\end{equation}
where the loss is normalized by the time span $T$.

\subsection*{Experimental setup}

\subsubsection*{Graph topologies}
We use two types of graphs for our experiments: synthetic random graphs and real-world graphs. 

The synthetic random graphs are generated using four common models: 
the \ac{er} \cite{erdHos1960evolution},
\ac{ba} \cite{barabasi1999emergence},
\ac{ws} \cite{watts1998collective} and
\ac{rgg} \cite{penrose2003random}.
We generate instances of \ac{er}, \ac{ba}, \ac{ws}, and \ac{rgg} graphs using different random seeds, and for each graph model, we average over the graph evaluations to obtain our final results.

For the real-world graphs, we use contiguous region graphs, cell phone mobility graphs, plane graphs, and other transportation graphs.
For the contiguous region graphs, the nodes are geometric administrative units such as countries, states, or cities, where two nodes establish a link if they share a border. 
We examine contiguous region graphs for the US, EU, China, and Africa  \cite{yu2023spatio, nr, githubGitHubP1seccountry_adjacency, gitGeochina}. 
The cell phone mobility graphs are built by collecting cell phone mobility data representing the proportion of people commuting between two nodes.
We examine weighted cell phone mobility graphs from Germany and the US \cite{kang2020multiscale, osfCovid19Mobility}.
For plane graphs representing the commute between two airports, we create a connected graph using the top $b$ busy airlines. By adjusting $b$, we can obtain graphs of varying sizes. We consider plane graphs both globally and within the US \cite{gitAirportnetworkdata, btsOST_RTranstats}.
For other transportation graphs, we include the bus, car, plane, and train graphs from the Spanish transport network\cite{spainTrans2024}. 

To generate the mobility matrix, we assign a $MobilityRate$ value corresponding to the fraction of the population flow to each link.
For simplicity, we assign the same value to each link in the synthetic random graphs and unweighted real graphs, which represent the mobility rate.
Finally, note that cell phone mobility graphs are weighted graphs, and therefore, there is no need to assign weights to them. In Appendix~\nameref{subsec:realgraphattribs} we provide further details on the real graphs used in this paper, including their graph attributions. 
All experiments were conducted using our open-source implementation, available at \href{https://github.com/bizhili/epi_fit_2024}{GitHub Repository}.

\subsubsection*{Evaluation metrics}
The following metrics were used to evaluate our topology inference performance:
\begin{enumerate}
    \item Spectral Similarity \cite{gera2018identifying}:
$$\frac{ \lambda(A)\cdot \lambda(\hat{A})}{ \|\lambda(A)\|_2  \|\lambda(\hat A)\|_2 }$$
where $\lambda(A)$  are the eigenvalues of matrix $A$, and $\hat{A}$ is the learned mobility matrix. The spectral similarity takes the eigenvalues into account as a graph embedding, highlighting the similarity of overall structure and connectivity \cite{abdi2021regular}.
    \item Pearson Correlation:
    $$\frac{cov(A,\hat{A})}{std(A)\cdot std(\hat{A})}$$
where $cov(A,\hat{A})$ represents the covariance of two matrices, and $std(A)$ is the standard deviation of matrix $A$. 
The Pearson correlation directly computes the specific link-to-link correlation. 
 
    \item Jaccard Similarity:
    $$\frac{\sum{min(A, \hat{A})}}{\sum{max(A, \hat{A})}}$$
    The Jaccard similarity measures the intersection divided by the union of the edge sets. 
    \item PR-AUC: The area under the precision-recall curve. The PR-AUC represents the precision-recall trade-off, and it is especially useful when dealing with sparse graphs.
\end{enumerate}


Finally, to evaluate the accuracy of our epidemic parameter estimations, we use the \ac{rmse} of $\hat{\beta},\hat{\gamma}$.

\subsubsection*{Parameter settings}
\phantomsection
\addcontentsline{toc}{section}{Acknowledgments}
\label{subsubsec: parameter_settings}

\begin{table}[ht]
\centering
\begin{tabular}{|c|c|}
\hline
Name  & Possible values \\
\hline
$Graph$ & \ac{er}, \ac{ba}, \ac{ws}, \ac{rgg}, or real graphs\\
\hline
$RandomSeed$& Integer\\
\hline
$n$& Positive integer\\
\hline
$Number\ of\ Pathogen$& Positive integer \\
\hline
$DLModel$& \acs{dtef}, \acs{ftef}, Infer2018\\
\hline
$Dense$& Positive integer\\
\hline
$MobilityRate$& Float in the range 0-1 \\
\hline
\end{tabular}

\caption{Configurable parameter settings for training.}
\label{tab: changeable_parameters}
\end{table}
Table~\ref{tab: changeable_parameters} lists the configurable settings for training: 
 the $Graph$ parameter is the synthetic random graph model or real graph used to perform simulations and make predictions;
 $RandomSeed$ is used to set random seeds for generating different graphs; $n$ is the number of nodes; 
 $Pathogen$ is the number of simulated pathogens; $Dense$ controls the average degree of the synthetic random graphs; and $MobilityRate$ is the percentage of the population traveling from one subpopulation to another in a unit of time.
By default, we simulate 20 synthetic random graphs (five graphs for each random graph model: \ac{er}, \ac{ba}, \ac{ws}, \ac{rgg}), with $n$=100 nodes per graph; then for each graph, we collect data from four pathogens, where the seed nodes are selected with probability proportional to node degree, reflecting the realistic scenario where initial infections are more likely to occur in highly connected regions. The average degree and mobility rate are set to $Dense$=4 and $MobilityRate$=0.01, respectively, unless otherwise specified.

The $DLModel$ parameter is the chosen \ac{dl} model. 
By combining different topology inference and epidemic learning configurations, we  obtain the following two \ac{dl} models\footnote{For brevity, we omit the \ac{esc} model, as it did not yield meaningful results.}:
\begin{enumerate}
    \item \Acf{dtef} (Eqs.~\eqref{eq: inference_model_all} and ~\eqref{eq: process_all}). 
    \item \Ac{ftef} (Eqs.~\eqref{eq: fast topology inference} and ~\eqref{eq: process_all}).
\end{enumerate}
We also include one competitive \ac{dl} model, Infer2018 \cite{wang2018inferring} for comparison. 



For the \ac{ws} graphs, we set the probability of rewiring each edge to be $0.1$ for small-world phenomenon. The \ac{rgg} graphs are generated by scattering nodes on the 2-D plane with uniform distribution. We run epidemics on the graphs using up to four infectious pathogens in each simulation and set the mean values of the epidemic parameters to $\beta$=1.1 and $\gamma=\frac{1}{7.5}$ \cite{murphy2021deep}; their variance is set to $0.1$.

\section*{Results}

\label{sec:experiments}



\subsection*{Model convergence}

\begin{figure}[ht]
\centering
\includegraphics[width=1.0\textwidth]{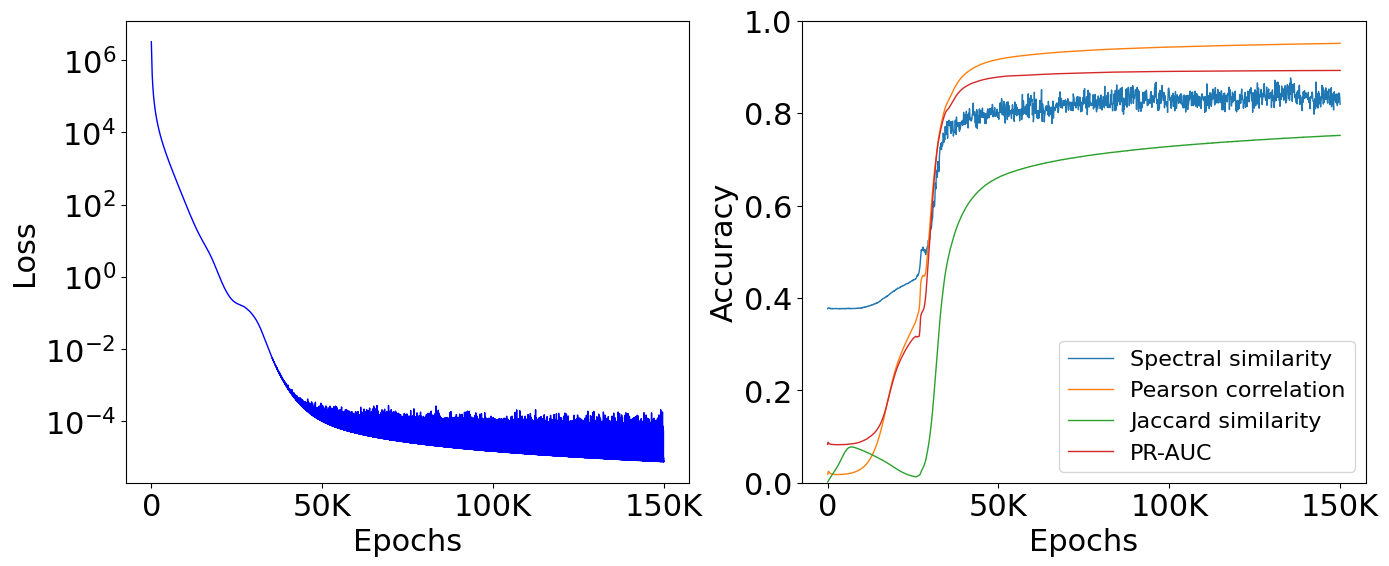}
\caption{The loss (Eq.~\eqref{eq: loss}) as a function of the number of optimization epochs (left); four evaluation metrics as a function of the number of optimization epochs (right).}
\label{fig: loss}
\end{figure}
Our \ac{dl} model's training process involved optimizing its parameters, using infectious data to learn the graph topology.
As seen in the graph on the left side of Figure~\ref{fig: loss}, the training loss consistently decreased as the number of optimization steps increased, demonstrating successful prediction of the node's infectious data. 
In the graph on the right side of Figure~\ref{fig: loss}, four evaluation metrics comparing the real and inferred adjacency also consistently improved, suggesting the highest extraction of topological information from the node's infectious data. 
Thus, our proposed model converges effectively to infer the right links. 
An adjacency matrix comparison of the ground-truth and inferred RGG graphs is provided in  Appendix~\nameref{subsubsec:adjcomp}. 
The fluctuations of the loss and spectral similarity (see Figure~\ref{fig: loss}) are predominantly due to the sparsity of the inferred mobility matrix.
Specifically, as the model converges, the inferred mobility matrix becomes more sparse, resulting in the observed fluctuations. In Appendix~\nameref{subsubsec:sparseidx} we validate this by computing the sparsity index \cite{goswami2021sparsity} of the inferred mobility matrix as a function of the optimization epochs. 

\subsection*{Benchmark results }
\label{sec:result:accuracy}

\begin{table}[!htbp]
\caption{Topology reconstruction benchmarks on representative random and real graphs.
This experiment is performed using four pathogens.
We show the performance of the five fitting models and a random Baseline (RND), based on four evaluation metrics as described in the text. The absence of PR-AUC values for Mobility US is due to them being weighted graphs. The bottom-line win rate represents the percentage of tasks where the model achieves a higher accuracy score than the other models.}
\begin{adjustwidth}{-0.5in}{0in}
\begin{tabular}{llHlHlllHlHll}
\hline
\hline
\multicolumn{1}{c}{} & \multicolumn{6}{|c}{Spectral similarity} & \multicolumn{6}{|c}{Pearson correlation} \\ \hline
 & DTEF & DTES & FTEF & FTES & Infer2018 & RND & DTEF & DTES & FTEF & FTES & Infer2018 & RND \\
 \hline
ER & 0.17 & \textbf{0.32} & \textbf{0.26} & 0.14 & 0.11 & 0.03 & \textbf{0.62} & 0.02 & 0.02 & 0.02 & 0.03 & 0 \\
BA & 0.22 & 0.4 & { \textbf{0.31}} & 0.21 & 0.14 & 0.05 & \textbf{0.72} & 0.04 & 0.03 & 0.03 & 0.03 & 0 \\
WS & \textbf{0.74} & 0.29 & { 0.55} & 0.14 & 0.1 & 0.1 & \textbf{0.88} & 0.02 & 0.14 & 0.13 & 0.12 & 0 \\
RGG & \textbf{0.82} & 0.42 & { 0.62} & 0.41 & 0.44 & 0.09 & \textbf{0.95} & 0.14 & 0.18 & 0.24 & 0.22 & 0 \\
Contiguous US & \textbf{0.8} & 0.01 & { 0.55} & 0.64 & 0.4 & 0.01 & \textbf{0.98} & 0.01 & 0.14 & 0.19 & 0.17 & 0.01 \\
Contiguous China&\textbf{0.98}&0.5&0.77&0.65&0.8&0.41&   \textbf{1.0}&0.26&0.09&0.24&0.09&0.02 \\
Contiguous EU&0.58&0.54&\textbf{0.59}&0.45&0.25&0.02& \textbf{1.0}&0.17&0.98&0.32&0.2&0.13\\
Contiguous Africa&\textbf{0.85}&0.44&0.67&0.64&0.35&0.09&   \textbf{0.99}&0.2&0.18&0.27&0.17&0.02\\
Mobility Germany&0.04&0.01&\textbf{0.14}&\textbf{0.25}&\textbf{0.14}&0.01&   \textbf{0.54}&0&0.04&0.05&0.02&0\\
Mobility US & \textbf{0.83} & 0.15 & { 0.38} & 0.39 & 0.44 & 0.04 & \textbf{0.96} & 0.02 & 0.02 & 0.04 & 0.06 & 0 \\
Global plane & 0.05 & 0.08 & \textbf{ 0.1} & \textbf{0.11} & 0.01 & 0.05 & \textbf{0.82} & 0.01 & 0.08 & 0.05 & 0.1 & 0 \\
US plane&0.53&0.06&0.64&0.29&\textbf{0.65}&0.34&   \textbf{0.71}&0.01&0.09&0.08&0.04&0\\
Spanish bus&\textbf{0.64}&0.62&0.49&0.42&0.43&0.49&   \textbf{0.92}&0.05&0.08&0.43&0.11&0.02 \\
Spanish car & 0.49 & 0.22 & { \textbf{0.61}} & 0.39 & 0.44 & 0.06 & \textbf{0.91} & 0 & 0.14 & 0.21 & 0.09 & 0.01 \\
Spanish plane&\textbf{0.72}&0.76&0.21&\textbf{0.86}&0.36&0.49&   \textbf{0.92}&0.29&0.01&0.7&0.04&0.03\\   
Spanish train & 0.7 & 0.55 & { \textbf{0.8}} & 0.26 & 0.63 & 0.24 & \textbf{0.84} & 0.09 & 0.07 & 0.17 & 0.06 & 0 \\
 \textbf{Win rate}& \textbf{0.5} & 0 & \textbf{0.5} & 0 & 0.13 & 0 & \textbf{1} & 0 & 0 & 0 & 0 & 0 \\

\hline
\multicolumn{1}{c}{} & \multicolumn{6}{|c}{Jaccard  similarity} & \multicolumn{6}{|c}{PR-AUC} \\
\hline
 & DTEF & DTES & FTEF & FTES & Infer2018 & RND & DTEF & DTES & FTEF & FTES & Infer2018 & RND \\
 \hline
ER & \textbf{0.25} & 0.02 & 0 & 0.01 & 0.01 & 0 & \textbf{0.57} & 0.08 & 0.09 & 0.1 & 0.1 & 0.12 \\
BA & \textbf{0.34} & 0.02 & 0 & 0.02 & 0.02 & 0 & \textbf{0.69} & 0.09 & 0.09 & 0.09 & 0.09 & 0.11 \\
WS & \textbf{0.58} & 0.02 & 0 & 0.02 & 0.02 & 0 & \textbf{0.86} & 0.08 & 0.13 & 0.12 & 0.12 & 0.12 \\
RGG & \textbf{0.75} & 0.01 & 0 & 0.03 & 0.03 & 0 & \textbf{0.89} & 0.12 & 0.15 & 0.17 & 0.17 & 0.12 \\
Contiguous US & \textbf{0.83} & 0 & 0 & 0.03 & 0.02 & 0 & \textbf{0.83} & 0.15 & 0.14 & 0.19 & 0.15 & 0.02 \\
Contiguous China&\textbf{0.96}&0.02&0.01&0.08&0.01&0&   \textbf{0.83}&0.23&0.19&0.3&0.19&0.22 \\
Contiguous EU&\textbf{0.96}&0.02&0.89&0.06&0.04&0&\textbf{0.8}&0.19&0.79&0.33&0.31&0.31\\
Contiguous Africa&\textbf{0.89}&0.02&0&0.04&0&0&   \textbf{0.83}&0.15&0.18&0.22&0.16&0.16\\
Mobility Germany&\textbf{0.22}&0&0&0&0&0&   \textbackslash{} & \textbackslash{} & \textbackslash{} & \textbackslash{} & \textbackslash{} & \textbackslash{}\\   
Mobility US & \textbf{0.48} & 0 & 0 & 0 & 0 & 0 & \textbackslash{} & \textbackslash{} & \textbackslash{} & \textbackslash{} & \textbackslash{} & \textbackslash{} \\
Global plane & \textbf{0.46} & 0 & 0 & 0.01 & 0.01 & 0 & \textbf{0.79} & 0.05 & 0.08 & 0.05 & 0.07 & 0.02 \\ 
US plane&\textbf{0.35}&0&0&0.01&0&0&   \textbf{0.68}&0.05&0.08&0.07&0.06&0.06\\
Spanish bus&\textbf{0.69}&0.02&0.01&0.2&0.01&0&   \textbf{0.82}&0.1&0.14&0.41&0.16&0.17 \\
Spanish car & \textbf{0.64} & 0 & 0 & 0.07 & 0.02 & 0 & \textbf{0.81} & 0.15 & 0.13 & 0.21 & 0.11 & 0.04 \\
 Spanish plane&\textbf{0.73}&0.03&0.02&0.6&0.03&0&   \textbf{0.85}&0.28&0.21&0.79&0.28&0.22\\ 
Spanish train & \textbf{0.52} & 0 & 0.01 & 0.06 & 0.02 & 0 & \textbf{0.76} & 0.22 & 0.12 & 0.16 & 0.1 & 0.03 \\ 
 \textbf{Win rate}& \textbf{1} & 0 & 0 & 0 & 0 & 0 & \textbf{1} & 0 & 0 & 0 & 0 & 0 \\
\hline
\hline
\end{tabular}
\label{tab: benchmark}
\end{adjustwidth}
\end{table}

\begin{table}[!htbp]
\caption{Topology reconstruction benchmarks on representative random and real graphs with epidemic parameters set to their ground truth values. Other settings are the same as \cref{tab: benchmark}.}
\begin{adjustwidth}{-0.5in}{0in}
\begin{tabular}{llHlHlllHlHll}
\hline
\hline
\multicolumn{1}{c}{} & \multicolumn{6}{|c}{Spectral similarity} & \multicolumn{6}{|c}{Pearson correlation} \\ \hline
 & DTEF & DTES & FTEF & FTES & Infer2018 & RND & DTEF & DTES & FTEF & FTES & Infer2018 & RND \\
 \hline
ER&0.16&\textbf{0.32}&\textbf{0.28}&0.15&0.09&0.03&0.61&0.01&\textbf{0.65}&0.03&0.57&0\\
BA&0.14&\textbf{0.41}&0.06&0.39&\textbf{0.41}&0.05&0.75&0.02&\textbf{0.77}&0.0&0.66&0\\
WS&\textbf{0.8}&0.29&0.71&0.12&0.75&0.1&0.92&0.01&\textbf{0.94}&0.02&0.91&0\\
RGG&0.8&0.39&\textbf{0.83}&0.19&0.76&0.09&\textbf{0.97}&0.01&0.93&0.03&0.85&0\\
Contiguous US&\textbf{0.97}&0.35&0.9&0.13&0.83&0.01&0.99&0.01&\textbf{1.0}&0.14&0.98&0.01\\
Contiguous China&0.68&0.44&\textbf{0.82}&0.25&0.78&0.41&\textbf{1.0}&0.03&\textbf{1.0}&0.4&0.99&0.02\\
Contiguous EU&0.49&0.5&\textbf{0.6}&0.22&0.43&0.02&\textbf{1.0}&0.05&\textbf{1.0}&0.26&0.96&0.13\\
Contiguous Africa&0.84&0.37&\textbf{0.86}&0.29&0.68&0.09&\textbf{0.98}&0.02&\textbf{0.98}&0.23&0.92&0.02\\
Mobility Germany&\textbf{0.13}&0.12&0.01&0.06&0.05&0.01&\textbf{0.37}&0.02&0.36&0.0&0.06&0\\
Mobility US&0.16&\textbf{0.4}&0.11&0.11&\textbf{0.25}&0.04&0.29&0.04&\textbf{0.74}&0.18&0.0&0\\
Global Plane &\textbf{0.07}&0.0&0.05&0.02&0.04&0.05&\textbf{0.79}&0.02&0.78&0.02&0.61&0\\
US plane&0.61&\textbf{0.64}&0.57&0.62&\textbf{0.63}&0.34&\textbf{0.72}&0.15&0.55&0.01&0.54&0\\
Spanish bus&0.72&0.63&\textbf{0.79}&\textbf{0.84}&0.66&0.49&0.74&0.06&\textbf{0.77}&0.17&0.44&0.02\\
Spanish car&0.37&0.42&\textbf{0.63}&0.53&0.34&0.06&\textbf{0.92}&0.04&0.47&0.38&0.78&0.01\\
Spanish plane&0.76&0.8&\textbf{0.86}&0.43&0.69&0.49&\textbf{0.91}&0.09&0.89&0.45&0.74&0.03\\
Spanish train&\textbf{0.71}&0.64&0.59&0.5&0.67&0.24&\textbf{0.87}&0.03&0.85&0.29&0.54&0\\
 \textbf{Win rate}& 0.31 & 0 & \textbf{0.56} & 0 & 0.19 & 0 & \textbf{0.63} & 0 & 0.56 & 0 & 0 & 0 \\
\hline
\multicolumn{1}{c}{} & \multicolumn{6}{|c}{Jaccard  similarity} & \multicolumn{6}{|c}{PR-AUC} \\
\hline
 & DTEF & DTES & FTEF & FTES & Infer2018 & RND & DTEF & DTES & FTEF & FTES & Infer2018 & RND \\
 \hline
ER&0.26&0.01&\textbf{0.34}&0.03&0.32&0&0.56&0.08&\textbf{0.61}&0.09&0.52&0.12\\
BA&0.39&0.01&\textbf{0.48}&0.01&0.44&0&0.73&0.08&\textbf{0.75}&0.08&0.67&0.11\\
WS&0.68&0.01&\textbf{0.78}&0.01&0.75&0&\textbf{0.88}&0.08&\textbf{0.88}&0.08&0.87&0.12\\
RGG&\textbf{0.81}&0.01&0.76&0.01&0.68&0&\textbf{0.9}&0.08&0.89&0.08&0.89&0.12\\
Contiguous US&0.92&0.02&\textbf{0.96}&0.08&0.93&0&\textbf{0.83}&0.09&\textbf{0.83}&0.23&0.82&0.02\\
Contiguous China&0.93&0.03&\textbf{0.96}&0.22&\textbf{0.96}&0&\textbf{0.83}&0.12&\textbf{0.83}&0.49&\textbf{0.83}&0.22\\
Contiguous EU&0.94&0.02&\textbf{0.95}&0.17&0.89&0&\textbf{0.8}&0.14&\textbf{0.8}&0.37&0.78&0.31\\
Contiguous Africa&0.86&0.02&\textbf{0.87}&0.1&0.83&0&\textbf{0.83}&0.1&\textbf{0.83}&0.28&0.8&0.16\\
Mobility Germany&\textbf{0.15}&0.0&0.14&0.0&0.02&0&\textbackslash{}&\textbackslash{}&\textbackslash{}&\textbackslash{}&\textbackslash{}&\textbackslash{}\\
Mobility US&\textbf{0.33}&0.0&0.32&0.01&0.0&0&\textbackslash{}&\textbackslash{}&\textbackslash{}&\textbackslash{}&\textbackslash{}&\textbackslash{}\\
Global Plane &0.43&0.01&\textbf{0.48}&0.0&0.43&0&\textbf{0.75}&0.05&\textbf{0.75}&0.04&0.7&0.02\\
US plane&\textbf{0.38}&0.01&0.36&0.01&0.36&0&\textbf{0.72}&0.03&0.66&0.05&0.65&0.06\\
Spanish bus&0.42&0.02&\textbf{0.48}&0.13&0.33&0&0.68&0.09&\textbf{0.7}&0.27&0.54&0.17\\
Spanish car&\textbf{0.68}&0.02&0.42&0.2&0.65&0&\textbf{0.81}&0.1&0.65&0.35&0.74&0.04\\
Spanish plane&0.7&0.03&\textbf{0.71}&0.33&0.58&0&\textbf{0.85}&0.19&0.84&0.55&0.8&0.22\\
Spanish train&0.6&0.02&\textbf{0.61}&0.18&0.46&0&\textbf{0.78}&0.09&\textbf{0.78}&0.31&0.69&0.03\\
 \textbf{Win rate}& 0.31 & 0 & \textbf{0.69} & 0 & 0.06 & 0 & \textbf{0.79} & 0 & 0.71 & 0 & 0.07 & 0 \\
\hline
\hline
\end{tabular}
\label{tab: benchmark2}
\end{adjustwidth}
\end{table}

To evaluate the performance of our two proposed topology inference models: \ac{dtef} and \ac{ftef}, we conduct a comprehensive benchmark experiment on representative synthetic random graphs and real graphs. 
We compare the performance of our models to a state-of-the-art model, Infer2018 \cite{wang2018inferring}.
In this experiment, we also include a random baseline (called RND) for comparison, to confirm that the rest of the models are capturing meaningful information from the data rather than merely making random guesses.
The results of this experiment are presented in Table~\ref{tab: benchmark}. As can be seen, in the vast majority of cases, the \acs{dtef} model outperformed the other models.

For completeness, in \Cref{tab: benchmark2} we also perform an ablation study to evaluate the topology inference performance of our models given known epidemic parameters, by setting these parameters to their ground truth values. We find that both \ac{ftef} and \ac{dtef} produce reasonable results, with \ac{ftef} occasionally outperforming \ac{dtef} while infer2018 gives suboptimal results.
Overall, these results prove the effectiveness of \ac{dti} model, with our proposed \acs{dtef} model consistently capturing the underlying structure, particularly for the \ac{ws} and \ac{rgg} graph topologies. Notably, \acs{dtef} clearly outperforms other models in more intricate real-world graph scenarios.
Further \ac{dtef} inference results on real graphs are presented in Appendix~\nameref{subsubsec:topinfrealres}.

\begin{table}[!htbp]
\caption{Epidemic parameter estimations of $\hat{\beta}$ (upper Table) and $\hat{\gamma}$ (lower Table) on both synthetic random and real-world graphs.
The performance of our different model configurations is evaluated based on the \ac{rmse} metric. In addition, the results of Infer2018, Bayes and NLS are shown for comparison (see text).} 
\begin{adjustwidth}{-1.7in}{1in}
\begin{tabular}{llHlHllllHlHlll}
\hline
\hline
&\multicolumn{7}{|c}{RMSE($\beta$)}& \multicolumn{7}{|c}{RMSE(1/$\gamma$)}\\
\hline
 |& DTEF & DTES & FTEF & FTES & Infer2018 & Bayes& NLS &DTEF & DTES & FTEF & FTES & Infer2018 & Bayes& NLS \\
 \hline
ER & \textbf{0} & 0.65 & { 1.05} & 0.64 & 0.69 &0.02 &0.01& {0.01} & 15.75 & 1.79 & 1.7 & 3.01 &1.49 &\textbf{0}\\
BA & \textbf{0} & 0.65 & { 1.04} & 0.55 & 0.55 &0.02 &0.01& {0.02} & 15.75 & 1.84 & 2.01 & 2.76 &1.30 &\textbf{0}\\
WS & \textbf{0} & 0.66 & { 1.01} & 0.64 & 0.62 &0.02 &0.03& {0.01} & 15.83 & 0.4 & 3.3 & 2.57 &1.79 &\textbf{0}\\
RGG & \textbf{0} & 0.68 & { 0.97} & 0.54 & 0.57 &0.02 &0.01& {0.02} & 14.38 & 0.5 & 6.58 & 6.84 &1.30 &\textbf{0}\\
Contiguous US & \textbf{0} & 0.2 & { 0.16} & 0.07 & 0.21 &\textbf{0}  & 0.01&\textbf{0} & 6.04 & 0.12 & 0.61 & 6.93&0.47 &\textbf{0}\\
Contiguous China & \textbf{0} & 0.2 &  0.17 & 0.05 & 0.17  &\textbf{0} & 0.01&\textbf{0} & 7.74 &  0.15 & 1.56 & 0.3 &0.52 &\textbf{0}\\
Contiguous EU & \textbf{0} & 0.19 &  \textbf{0} & 0.03 & 0.03  &0.01 &0.01& \textbf{0} & 7.92 &  \textbf{0} & 1.01 & 0.23 &1.04 &\textbf{0}\\
Contiguous Africa & \textbf{0} & 0.2 &  0.16 & 0.06 & 0.18  &0.01 &\textbf{0}& \textbf{0} & 6.95 &  0.26 & 0.48 & 0.63 &1.08 &\textbf{0}\\
Mobility Germany & \textbf{0} & 1.12 & 0.22 & 0.22 & 0.22  &\textbf{0} &0.26& \textbf{0} & 6.2 & 0.67 & 0.63 & 0.77&0.12 &0.12\\
Mobility US & \textbf{0} & 0.2 & { 0.18} & 0.07 & 0.2 &\textbf{0} &0.74& \textbf{0} & 6.63 & 0.61 & 0.21 & 6.74 &0.08 &0.22\\
Global plane & \textbf{0} & 1.12 & { 0.21} & 0.17 & 0.22 &\textbf{0} &\textbf{0}& {0.01} & 6.3 & 0.3 & 0.38 & 4.1 &0.45&\textbf{0}\\
US plane & \textbf{0} & 1.12 & 0.22 & 0.25 & 0.22  &\textbf{0} &\textbf{0}& {0.03} & 6.32 & 0.55 & 0.91 & 0.53 &0.55&\textbf{0}\\
Spanish bus & \textbf{0} & 1.12 & { 0.22} & 0.24 & 0.22 &\textbf{0}&0.01& \textbf{0} & 5.39 &  0.37 & 0.2 & 0.24&0.45&\textbf{0}\\
Spanish car & \textbf{0} & 0.2 & { 0.18} & 0.02 & 0.2 &0.01 &\textbf{0}& \textbf{0} & 7.03 & 0.16 & 0.7 & 7.71&1.08&\textbf{0}\\
Spanish plane & \textbf{0} & 0.18 & { 0.12} & \textbf{0} & 0.09 &0.01 &\textbf{0}& \textbf{0} & 7.75 &  0.38 & \textbf{0} & 0.38&0.34&\textbf{0}\\
Spanish train & \textbf{0} & 0.2 & { 0.15} & 0.05 & 0.19 &0.01 &0.01& \textbf{0} & 6.37 & 3.56 & 0.34 & 7.25&0.34&\textbf{0}\\
  \textbf{Win rate} & \textbf{1} & 0 & 0.06 & 0.06 & 0 &0.44 &0.31& 0.63 & 0 & 0.06 & 0.06 & 0 &0 &\textbf{0.88}\\

\hline
\hline
\end{tabular}
\label{tab: benchmark_epi}
\end{adjustwidth}
\end{table}

We also compare the accuracy of the epidemic parameter inference across the five \ac{dl} models, 
using the \ac{rmse} of $\hat{\beta},1/\hat{\gamma}$ as our evaluation metric.
We generated five random graphs for the \ac{er}, \ac{ba}, \ac{ws}, and \ac{rgg} random graph models, using four pathogens. 
For comparative analysis, we included traditional Bayesian inference \cite{o1999bayesian} and the nonlinear least squares (NLS) \cite{taghizadeh2022seir} as baselines.
As detailed in \cref{tab: benchmark_epi}, the epidemic parameter estimation results show that the \ac{dtef} achieves performance comparable to the NLS method.
Furthermore, the parameter convergence to zero \ac{rmse} in the experiments confirms the identifiability of $\beta$ and $\gamma$ under \cref{eq: process_all}.

\subsection*{Effect of multiple pathogens}


\begin{figure*}[!htbp]
\begin{adjustwidth}{-2.5in}{0in}
\centering
\includegraphics[width=1.4\textwidth]{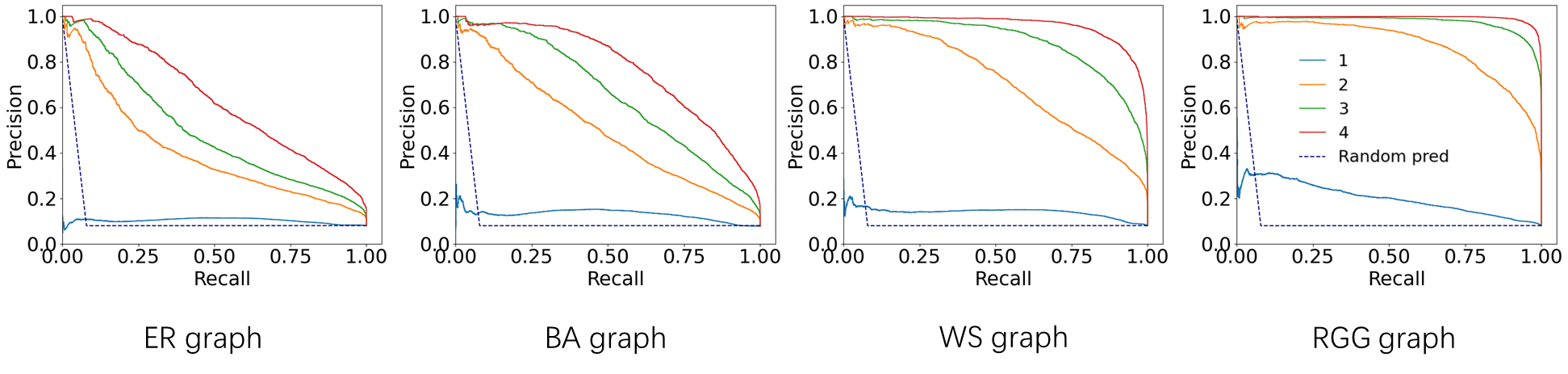}
\end{adjustwidth}
\caption{ DTEF model performance PR curves for various random graph models, with 1-4 pathogens and a random baseline.}
\label{fig: pr}
\end{figure*}

To further investigate the impact of incorporating data from multiple pathogens on topology inference accuracy, we conduct an experiment, in which data from 1-4 pathogens appears in each graph and DTEF serves as the DL model.
The rest of the parameters are set to their default settings (see subsection~\nameref{subsubsec: parameter_settings}).

Figure~\ref{fig: pr} presents the PR (precision recall) plot for different pathogens.
All of the PR curves are consistently above the baseline, indicating that the inferences surpass random chance in terms of discriminative power.
As the number of pathogens increases, the PR curves trend toward the top-right corner, indicating enhanced inference accuracy.
With one pathogen, the PR curve lies only slightly above random, because a single pathogen explores only a small subset of network paths.
Adding a second pathogen introduces a new seed node location and independent diffusion trajectory, making previously unobserved links identifiable.
After three or four pathogens, most mobility links are exposed, explaining the rapid improvement in PR-AUC and Jaccard similarity.

\begin{figure*}[ht!]
\begin{adjustwidth}{-2.5in}{0in}
\centering
\includegraphics[width=1.4\textwidth]{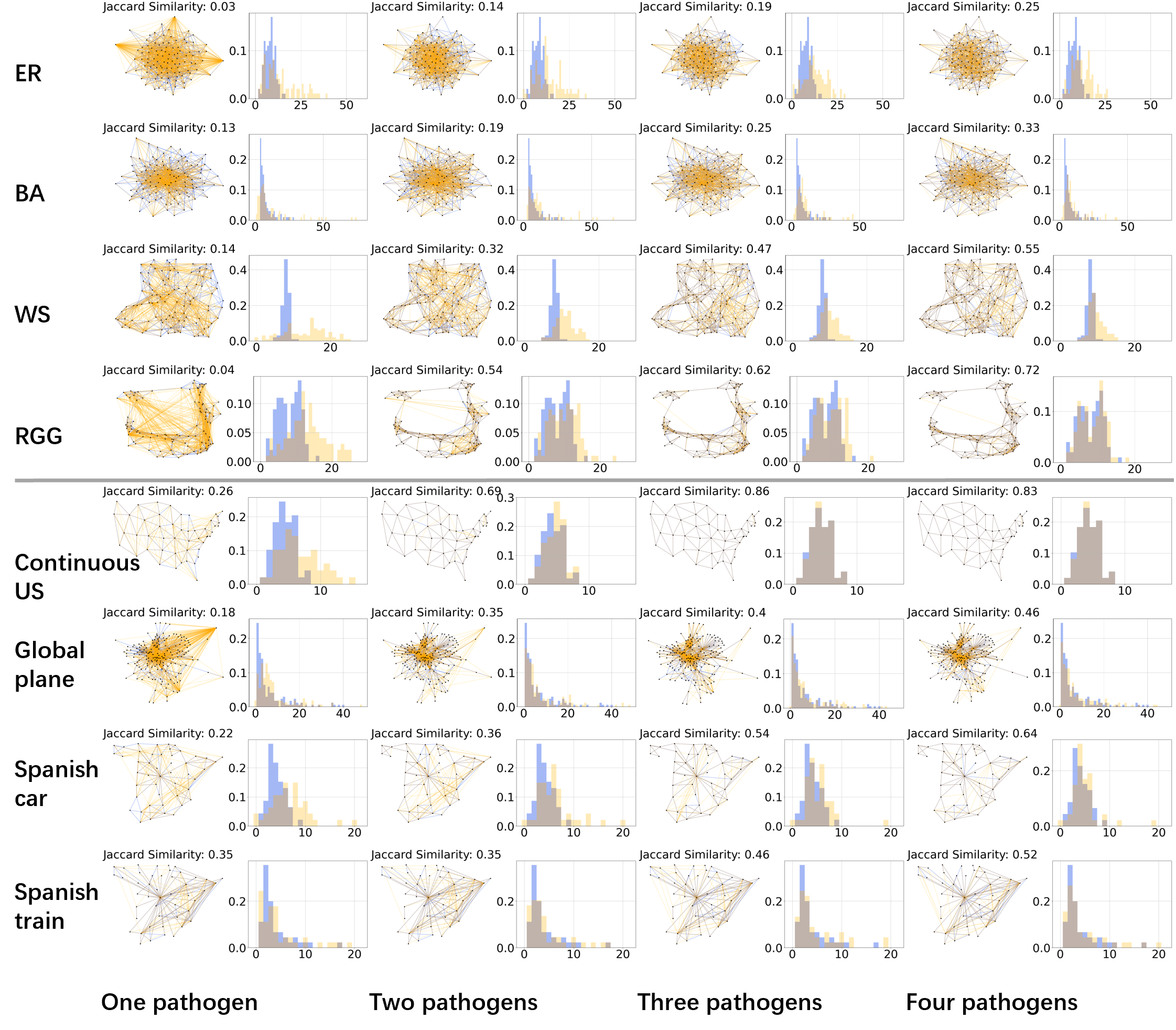}
\end{adjustwidth}
\caption{Evolution of the DTEF model's topology inference with different numbers of pathogens.
Each row contains four pairs of graphs: the spring layout of the graph titled with the Jaccard similarity where
the blue color represents the ground-truth graph, while the orange color is the inferred graph (left); the degree distribution histogram comparison with the x-label degree and y-label probability (right).}
\label{fig: pathogen_hst}
\end{figure*}

Figure~\ref{fig: pathogen_hst} offers a visual complement to our previous findings, depicting the spring layout and degree distribution of the evolving topology where the number of pathogens ranges from 1-4. 
For completeness, we also include the performance of the model over representative real graphs.
The predicted spring layouts progressively converge toward the ground-truth topology as the number of pathogens increases. 
This can also be observed in the degree distributions, where the degree distribution histogram of $\hat{A}$ aligns closely with $A$ starting from three pathogens, demonstrating that the model recovers both local connectivity and global graph statistics.
We find that the topology of real contiguous region graphs is almost entirely inferred, in agreement with our benchmark simulated results.
Therefore, multiple pathogens incorporate valuable information for topology inference, in that links connecting prior and later infected regions have a significant effect on the epidemic spread.

\subsection*{Effect of mobility rate}

\begin{figure*}[!htbp]
\begin{adjustwidth}{-2.5in}{0in}
\centering
\includegraphics[width=1.4\textwidth]{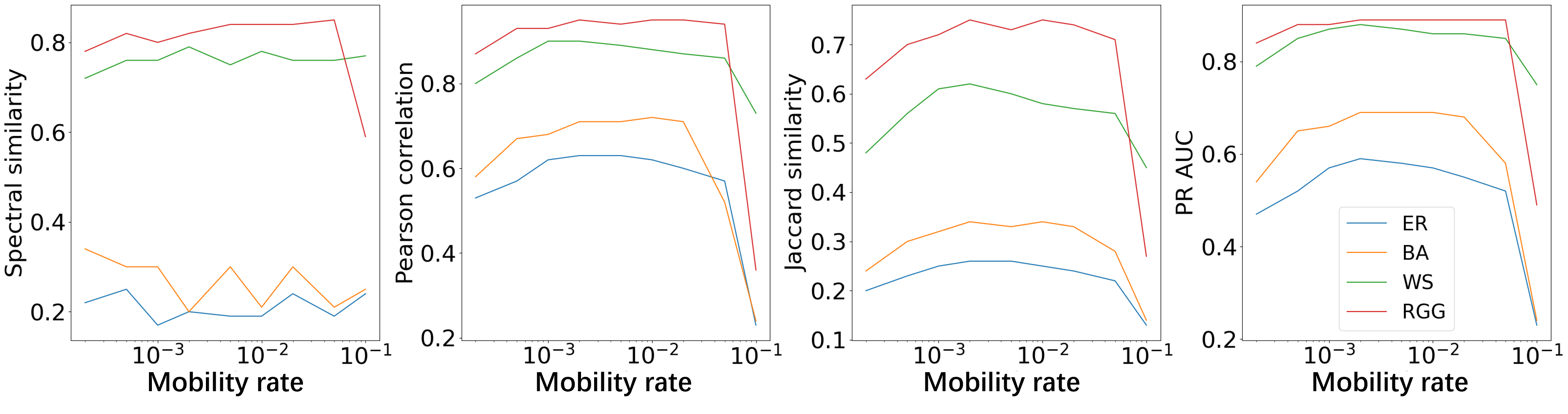}
\end{adjustwidth}
\caption{Effect of different mobility rates on the topology inference accuracy of the DTEF model.}
\label{fig: rate_acc}
\end{figure*}

To understand the impact of the mobility rate on the DTEF model's topology inference accuracy, we generate
each graph with $n$=100 nodes with an average degree of $\text{log}(n)$ and systematically vary the mobility rate across a range of values (2e-4, 5e-4, 1e-3, 2e-3, 5e-3, 1e-2, 2e-2, 5e-2, and 1e-1). The rest of the parameters are set to their default settings. 

As can be seen in Figure~\ref{fig: rate_acc}, which presents the results, there is a notable peak in inference accuracy near a mobility rate of 1e-2 (the corresponding numerical values are listed in Table~\ref{tab: rate_acc}). This suggests that an optimal mobility rate exists for effective topology inference, i.e., for capturing meaningful infection patterns within the data.

The results of a systematic study of the impact of other parameters on the DTEF model's inference accuracy are presented in Appendices~\nameref{subsubsec:denseres} and are summarized as follows: 
Dense graphs diffusing faster cause information saturation with decreased accuracy(e.g., \ac{er}, avg degree $4\rightarrow8$: Pearson correlation drops $20\%$).
A higher node count makes the graph topology more difficult to infer.

\section*{Discussion}

The current understanding of mobility graphs in epidemic modeling often assumes a static topology within a certain time frame.
We leverage this assumption to exploit the rich information offered by multiple pathogen epidemic data gathered during the epidemic time span, thereby increasing the accuracy of model fitting.

\ac{dtef} consistently reconstructs complex graph topologies using only epidemic time-series data, outperforming existing baselines across synthetic and real-world graphs.
Convergence is stable and rapid due to the fast-forward-backward mechanism, which circumvents the obfuscated gradient problem typically encountered in traditional learning methods. Furthermore, incorporating multiple pathogens substantially improves structural identifiability.
The model recovers not only the degree distributions and epidemic parameters but also the individual links with low error.
These results confirm that epidemic diffusion information is key to infer large-scale mobility networks, offering new prospects for epidemic modeling beyond topology learning.

We also systematically investigated the impact of various factors on inference accuracy.
Of the four random graph models examined (\ac{er}, \ac{ba}, \ac{ws}, and \ac{rgg}), \ac{dtef} achieved the highest inference accuracy on \ac{rgg}, followed by \ac{ws}.
Notably, this pattern extends to real-world networks: the model achieved the highest reconstruction accuracy on contiguous-region graphs, which have spatial locality constraints, like \ac{rgg}.

As previously stated, incorporating data from more pathogens consistently improved the results.
This can be explained by the fact that different pathogens explore the same mobility graph with different seed nodes and epidemic parameters, which provides more non-correlated structural information on the topology.
In addition, an ablation study revealed that running the model with known epidemic parameters facilitates topology inference, confirming the previous results in the literature. 

The ability to jointly infer parameters and topology serves as a significant methodological proof of concept.
It demonstrates that, under ideal conditions, information embedded in infection time series is sufficient to recover hidden mobility structures.

Despite these promising results, several limitations remain. First, our model assumes a static mobility topology within the observation window. While this holds for short term analyses, real world mobility is dynamic, often fluctuating due to seasonal and behavioral factors, as well as non-pharmaceutical interventions (e.g., lockdown). 
Second, as with many \ac{dl} approaches, inference quality depends on the granularity and quality of the input data; extremely sparse or noisy surveillance data may hinder convergence of the encoder-decoder architecture.
Finally, our current framework assumes a specific epidemic diffusion mechanism. Topology inference might suffer from bias in cases where the assumed compartmental model misrepresents the underlying disease dynamics.

Future extensions with increased robustness to noise may serve as surrogate epidemic models in the absence of mobility data. 
Further research directions include: exploring cross-immunity effects between pathogens, taking into account data-collection latency/noise, and diversifying input graphs by sampling subgraphs. We plan to focus on leveraging real epidemic data to further validate and refine the proposed approach, potentially improving inference accuracy and applicability in future outbreaks.

\section*{Supporting information}

\subsection*{\label{subsec:theoretical_proofs}Theoretical proofs}

\renewcommand{\thefigure}{S\arabic{figure}}
\renewcommand{\thetable}{S\arabic{table}}
\setcounter{figure}{0}
\setcounter{table}{0}

\subsubsection*{\label{subsec:alpha approx}Proof 1: $\alpha$ approximation }
Here, we prove that the approximate form for $\alpha$ (see Eq.~\eqref{eq:alpha-app}) holds, i.e.,
\begin{equation}
\label{eq:app_eq1}    
1-\left(1-{\frac{\beta}{P}}\right)^{I\odot P}\approx1-e^{-{\beta}I}
\end{equation}
\textbf{proof:}

Since $P$ is usually much larger than $\beta$ and $I$, we can therefore utilize the definition of the exponential function: 
\begin{equation}
\label{eq:app_eq2}
e^x=\lim_{n\rightarrow \infty}{(1+\frac{x}{n})^n}
\end{equation}
where we let  $x=-\beta I ,n=I_i\cdot P_i$. Thus, we obtain:
\begin{equation}
\label{eq:app_eq3}
e^{-\beta I}=\lim_{I\odot P\rightarrow \infty}{(1-\frac{\beta}{ P})^{I\odot P}}
\end{equation}

\subsubsection*{\label{subsec:sparsity}Proof 2: Sparsity of infection matrix}
Here, we prove the sparsity of matrix $Z$ in Eq.~\eqref{eq:infection_matrix}.

Given:
\begin{equation}
\label{eq:app_eq4}
Z=\boldsymbol{I}+\frac{(\sum_{dim=1}A)^T\odot A\odot P}{(\sum_{dim=1}(A\odot P))^T}
\end{equation}

we need to prove that $Z$ is sparse.

\textbf{proof:}

If $i\neq j$:
\begin{equation}
\label{eq:app_eq5}
Z(i, j)= \frac{(\sum A(i, :))\cdot A(i,j) \cdot P(j)}{A(i, :)^T\cdot P}
\end{equation}

Since $A(i, :)^T\cdot P$, $P(j)$ and $\sum A(i, :)$ are positive, then $Z$ acquires the sparsity attribution from $A$.

\subsubsection*{\label{subsec:reversibility}Proof 3: 
Reversibility of infection matrix  }
Here, we prove that reversing Eq.~\eqref{eq:infection_matrix} yields
Eq.~\eqref{eq:infection_reverse}.

Given:
\begin{equation}
\label{eq:app_eq6}
Z=\boldsymbol{I}+\frac{(\sum_{dim=1}A)^T\odot A\odot P}{(\sum_{dim=1}(A\odot P))^T}
\end{equation}

we need to prove:
\begin{equation}
\label{eq:app_eq7}
A=\frac{(Z-\boldsymbol{I}) \odot(\sum_{dim=1}(Z-\boldsymbol{I}))^T}{P\odot (\sum_{dim=1}\frac{Z-\boldsymbol{I}}{P})^T}
\end{equation}
\textbf{proof:}

Let $Z_2=Z-\boldsymbol{I}$.

By definition (see Eq.~\eqref{eq:infection_matrix}) we have:
\begin{equation}
\label{eq:app_eq8}
Z_2(i, j)= \frac{(\sum A(i, :))\cdot A(i,j) \cdot P(j)}{A(i, :)^T\cdot P}
\end{equation}

which upon summation yields
\begin{equation}
\label{eq:app_eq9}
\sum Z_2(i, :)= \sum A(i, :)
\end{equation}
Taking into account that
\begin{equation}
\label{eq:app_eq10}
\frac{Z_2(i, :)}{P}= \frac{\sum A(i, :)\cdot A(i, :)}{A(i, :)^T\cdot P}
\end{equation}
as well as
\begin{equation}
\label{eq:app_eq11}
\sum \frac{Z_2(i, :)}{P}= \frac{(\sum A(i, :))^2}{A(i, :)^T\cdot P}
\end{equation}
we take the ratio of Eqs.~\eqref{eq:app_eq10}--\eqref{eq:app_eq11} such that
\begin{equation}
\label{eq:app_eq12}
\frac{\frac{Z_2(i, :)}{P}}{\sum \frac{Z_2(i, :)}{P}}=\frac{A(i,:)}{\sum A(i,:)}
\end{equation}
Thus, we conclude:
\begin{equation}
\label{eq:app_eq13}
A(i,:)=\frac{Z_2(i, :)\cdot \sum Z_2(i, :)}{P\cdot \sum \frac{Z_2(i, :)}{P}}
\end{equation}

\subsection*{\label{subsec:realgraphattribs} Real-world graph attributions}

Table~\ref{tab: real_attribution} lists the attributions of all real graphs used in the paper.

\begin{table*}[!htbp]
\begin{tabular}{cccc}
\hline
\hline
\multicolumn{1}{l}{} & \multicolumn{1}{l}{Nodes} & \multicolumn{1}{l}{Links} & Weighted \\ \hline
Contiguous US & 49 & 107 & no \\
Contiguous China & 33 & 69 & no \\
Contiguous EU & 24 & 38 & no \\
Contiguous Africa & 47 & 103 & no \\
Mobility German & 52 & - & yes \\
Mobility US & 401 & - & yes \\
Global plane& 174 & 635 & no \\
US plane & 313 & 2323 & no \\
Spanish bus & 32 & 202 & no \\
Spanish car & 46 & 205 & no \\
Spanish plane & 47 & 205 & no \\
Spanish train & 46 & 205 & no \\
\hline
\hline
\end{tabular}
\caption{Real graphs' attributions, including the node number, link number, and whether the graph is weighted or not.}
\label{tab: real_attribution}
\end{table*}

\subsection*{\label{subsec:supp exp results} Supplementary experimental results}

\subsubsection*{\label{subsubsec:adjcomp}Adjacency matrix comparison for \ac{rgg} graphs}

\begin{figure}[!htbp]
\centering
\includegraphics[width=0.8 \textwidth]{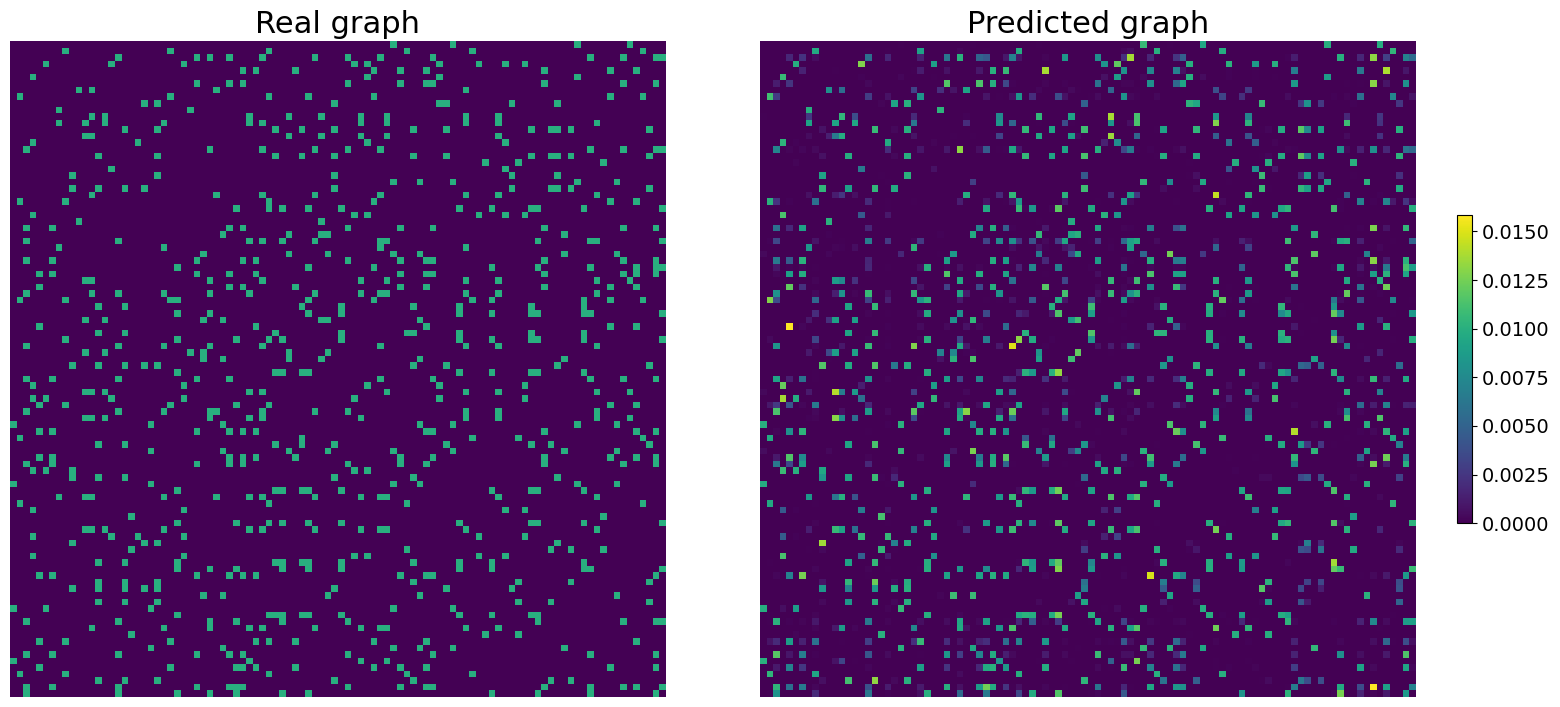}
\caption{Adjacency matrix comparison between ground-truth and inferred \ac{rgg} graphs containing 100 nodes with four pathogens: the real binary graph (left); the inferred adjacency topology according to the DTEF model with four pathogens of data (right).}
\label{fig: adjacency}
\end{figure}

To further illustrate the proposed model's robustness, Figure~\ref{fig: adjacency} presents a comparison between the ground-truth \ac{rgg} graph and the inferred \ac{rgg} graphs for scenarios with four distinct pathogens of data.
The left plot is the real binary graph, and the right plot is the inferred adjacency topology according to the DTEF model.
This comparison visually highlights the model's ability to learn and represent the structural relationships in the \ac{rgg}.

\subsubsection*{\label{subsubsec:sparseidx}Sparsity index}
\begin{figure*}[!htbp]
\centering
\includegraphics[width=0.6\textwidth]{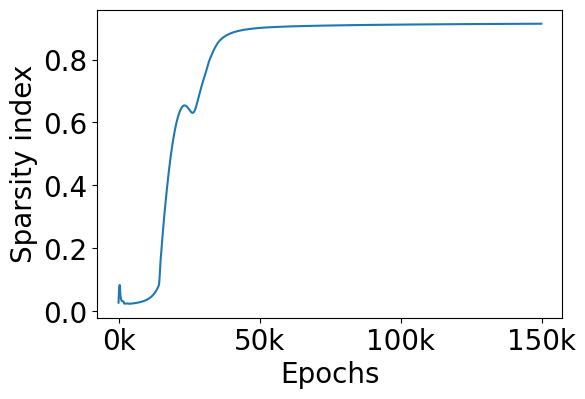}
\caption{Sparsity index of the mobility matrix as a function of the number of optimization epochs.}
\label{fig:sparsity}
\end{figure*}

The sparsity index of the mobility matrix as a function of the number of optimization epochs is presented in Figure~\ref{fig:sparsity}. 

\subsubsection*{\label{subsubsec:denseres}Effect of the average degree}

\begin{table*}[!htbp]
\begin{adjustwidth}{-2in}{0in}
\begin{tabular}{cccccccccccccccc}
\multicolumn{1}{l}{} & \multicolumn{1}{l}{} & \multicolumn{1}{l}{} & \multicolumn{1}{l}{} & \multicolumn{1}{l}{} & \multicolumn{1}{l}{} & \multicolumn{1}{l}{} & \multicolumn{1}{l}{} & \multicolumn{1}{l}{} & \multicolumn{1}{l}{} & \multicolumn{1}{l}{} & \multicolumn{1}{l}{} & \multicolumn{1}{l}{} & \multicolumn{1}{l}{} & \multicolumn{1}{l}{} & \multicolumn{1}{l}{} \\ \hline
\multicolumn{1}{c|}{} & \multicolumn{5}{c|}{Spectral   similarity} & \multicolumn{5}{c|}{Pearson correlation} & \multicolumn{5}{c}{Jaccard similarity} \\ \hline
\begin{tabular}[c]{@{}c@{}}Avg\\ degree\end{tabular} & 4 & 5 & 6 & 7 & 8 & 4 & 5 & 6 & 7 & 8 & 4 & 5 & 6 & 7 & 8 \\ \hline
\ac{er} & 0.2 & 0.22 & 0.32 & 0.19 & \textbf{0.33} & \textbf{0.61} & 0.56 & 0.52 & 0.51 & 0.49 & \textbf{0.25} & 0.23 & 0.21 & 0.22 & 0.22 \\
\ac{ba} & 0.21 & 0.38 & 0.43 & 0.36 & \textbf{0.48} & \textbf{0.72} & 0.63 & 0.64 & 0.6 & 0.59 & \textbf{0.34} & 0.28 & 0.29 & 0.28 & 0.27 \\
\ac{ws} & 0.78 & 0.75 & \textbf{0.83} & \textbf{0.83} & \textbf{0.83} & \textbf{0.88} & 0.87 & 0.84 & 0.82 & 0.79 & \textbf{0.59} & 0.57 & 0.54 & 0.52 & 0.5 \\
\ac{rgg} & 0.83 & 0.83 & 0.84 & 0.88 & \textbf{0.9} & 0.95 & \textbf{0.96} & 0.95 & 0.95 & 0.94 & 0.75 & \textbf{0.76} & \textbf{0.76} & 0.75 & 0.73 \\ \hline
\multicolumn{1}{c|}{}  & \multicolumn{5}{c|}{PR-AUC} &  &  &  &  & & & & &  \\ \hline
\begin{tabular}[c]{@{}c@{}}Avg\\ degree\end{tabular}  & 4 & 5 & 6 & 7 & 8 &  &  &  &  & &  &  &  &  &  \\ \hline
\ac{er} & \textbf{0.56} & 0.53 & 0.51 & 0.52 & 0.52 &  &  &  &  &  \\
\ac{ba} &  \textbf{0.69} & 0.62 & 0.65 & 0.62 & 0.62 &  &  &  &  & &  &  &  &  &  \\
\ac{ws} & 0.86 & \textbf{0.88} & 0.87 & 0.87 & 0.85 &  &  &  &  & &  &  &  &  &  \\
\ac{rgg}  & 0.89 & 0.92 & 0.93 & 0.94 & \textbf{0.95} &  &  &  &  & &  &  &  &  &  \\ \hline
\hline
\end{tabular}
\end{adjustwidth}
\caption{Effect of different average degrees on the topology inference accuracy of the DTEF model.}
\label{tab: dense_acc}
\end{table*}

\begin{figure*}[b]
\centering
\includegraphics[width=1\textwidth]{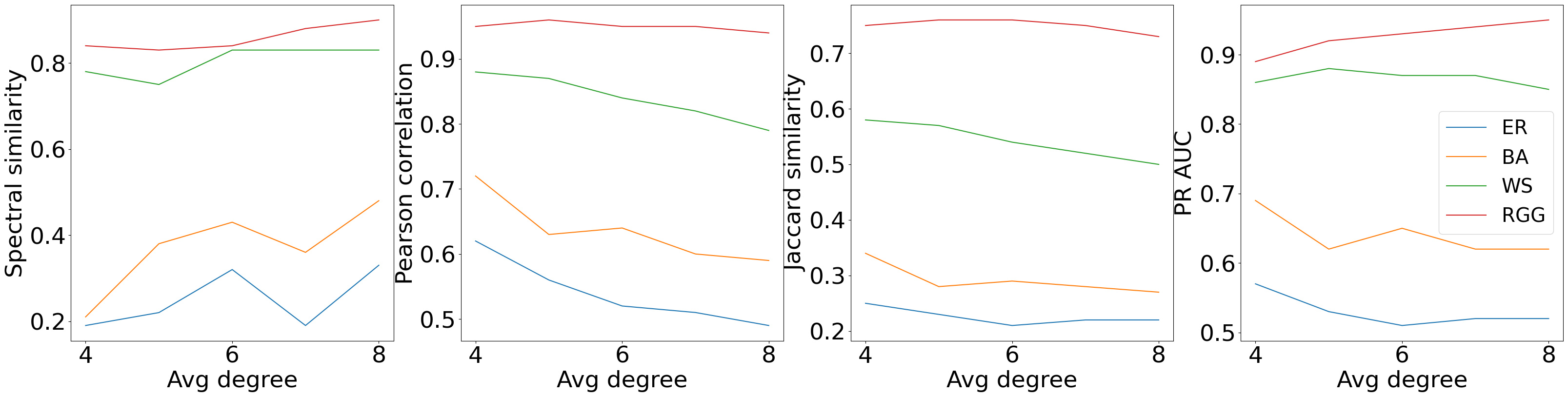}
\caption{Effect of different average degrees on the topology inference accuracy of the DTEF model.}
\label{fig: dense_acc}
\end{figure*}

To explore the relationship between the average degree, $Dense$, and the topology inference accuracy of the DTEF model, we simulate 100 synthetic random graphs at five different average degree values: 4, 5, 6, 7, and 8. The rest of the parameters are set at their default settings (see subsection~\nameref{subsubsec: parameter_settings}).

Figure~\ref{fig: dense_acc} presents the topology inference accuracy of the DTEF model for different average degree values (the corresponding numerical values appear in Table~\ref{tab: dense_acc}). 
As can be seen, increasing the average degree appears to negatively impact the model's ability to infer the graph topology.
This suggests that denser graphs pose a greater challenge to the \ac{dtef} model, potentially due to the larger number of potential connections masking the underlying structural relationships. 

\subsubsection*{Effect of the node number}

\begin{table*}[!htbp]
\begin{adjustwidth}{-2in}{0in}
\begin{tabular}{ccccccccccccccccc}
 \hline
 \hline
\multicolumn{1}{l|}{} & \multicolumn{4}{l|}{Spectral   similarity} & \multicolumn{4}{l|}{Pearson correlation} & \multicolumn{4}{l|}{Jaccard similarity} & \multicolumn{4}{l}{PR-AUC} \\ \hline
Nodes & \multicolumn{1}{r}{50} & \multicolumn{1}{r}{100} & \multicolumn{1}{r}{200} & \multicolumn{1}{r}{400} & \multicolumn{1}{r}{50} & \multicolumn{1}{r}{100} & \multicolumn{1}{r}{200} & \multicolumn{1}{r}{400} & \multicolumn{1}{r}{50} & \multicolumn{1}{r}{100} & \multicolumn{1}{r}{200} & \multicolumn{1}{r}{400} & \multicolumn{1}{r}{50} & \multicolumn{1}{r}{100} & \multicolumn{1}{r}{200} & \multicolumn{1}{r}{400} \\ \hline
\ac{er} & 0.18 & 0.2 & 0.12 & \textbf{0.21} & \textbf{0.85} & 0.61 & 0.46 & 0.35 & \textbf{0.52} & 0.25 & 0.14 & 0.07  & \textbf{0.82} & 0.56 & 0.37 & 0.22 \\
\ac{ba} & \textbf{0.45} & 0.21 & 0.12 & 0.22 & \textbf{0.87} & 0.72 & 0.55 & 0.49 & \textbf{0.56} & 0.34 & 0.19 & 0.13  & \textbf{0.82} & 0.69 & 0.48 & 0.38 \\
\ac{ws} & 0.76 & \textbf{0.78} & 0.76 & 0.68 & \textbf{0.97} & 0.88 & 0.78 & 0.73 & \textbf{0.81} & 0.59 & 0.41 & 0.34  & \textbf{0.87} & 0.86 & 0.78 & 0.71 \\
\ac{rgg} & \textbf{0.9} & 0.83 & 0.77 & 0.78 & \textbf{0.98} & 0.95 & 0.94 & 0.92 & \textbf{0.85} & 0.75 & 0.7 & 0.65 & 0.88 & 0.89 & \textbf{0.9} & 0.89\\
\hline
\hline
\end{tabular}
\end{adjustwidth}
\caption{Effect of different numbers of nodes on the topology inference accuracy of the DTEF model.}
\label{tab: n_acc}
\end{table*}

\begin{figure*}[!htbp]
\centering
\includegraphics[width=1\textwidth]{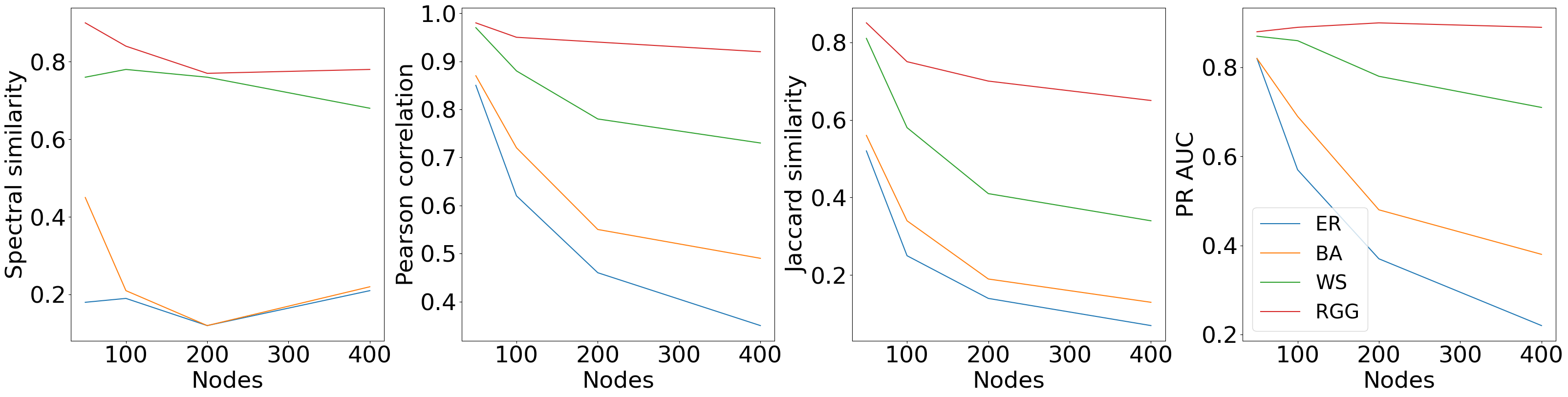}
\caption{Effect of different numbers of nodes on the topology inference accuracy of the DTEF model.}
\label{fig: n_acc}
\end{figure*}

To investigate the influence of the node count on the DTEF model's topology inference accuracy, we conduct an experiment involving 80 synthetic random graphs at four node count levels, $n$: 50, 100, 200, and 400. Correspondingly, the average node degree is set to $\text{log}(n)$, while the rest of the parameters are set to their default settings (see subsection~\nameref{subsubsec: parameter_settings}).

Figure~\ref{fig: n_acc} presents the topology inference accuracy results for the varying node counts (the corresponding numerical values appear in Table~\ref{tab: n_acc}).
Notably, the accuracy exhibits a consistent decline as the number of nodes increases.
This finding suggests that larger graphs present a greater challenge for accurate topology inference, potentially due to the amplified complexity of interactions and structural relationships within larger systems.

\subsubsection*{\label{subsubsec:mobires}Effect of the mobility rate}

\begin{table*}[!htbp]
\begin{adjustwidth}{-2.7in}{0in}
\begin{tabular}{ccccccccccccccccccc}
\hline \hline
\multicolumn{1}{c|}{} & \multicolumn{9}{c|}{Spectral   similarity} & \multicolumn{9}{c}{Pearson correlation} \\ \hline 
\begin{tabular}[c]{@{}c@{}}$M_R$\end{tabular} & 2e-4 & 5e-4 & 1e-3 & 2e-3 & 5e-3 & 1e-2 & 2e-2 & 5e-2 & 1e-1 & 2e-4 & 5e-4 & 1e-3 & 2e-3 & 5e-3 & 1e-2 & 2e-2 & 5e-2 & 1e-1 \\ \hline
\ac{er} & 0.22 & 0.25 & 0.17 & 0.2 & \textbf{0.19} & 0.2 & 0.24 & 0.19 & 0.24 & \textbf{0.53} & 0.57 & 0.62 & 0.63 & 0.63 & 0.61 & 0.6 & 0.57 & 0.23 \\
\ac{ba} & 0.34 & 0.3 & 0.3 & 0.2 & \textbf{0.3} & 0.21 & 0.3 & 0.21 & 0.25 & \textbf{0.58} & 0.67 & 0.68 & 0.71 & 0.71 & 0.72 & 0.71 & 0.52 & 0.24 \\
\ac{ws} & 0.72 & 0.76 & \textbf{0.76} & \textbf{0.79} & \textbf{0.75} & 0.78 & 0.76 & 0.76 & 0.77 & \textbf{0.8} & 0.86 & 0.9 & 0.9 & 0.89 & 0.88 & 0.87 & 0.86 & 0.73 \\
\ac{rgg} & 0.78 & 0.82 & 0.8 & 0.82 & \textbf{0.84} & 0.83 & 0.84 & 0.85 & 0.59 & 0.87 & \textbf{0.93} & 0.93 & 0.95 & 0.94 & 0.95 & 0.95 & 0.94 & 0.36 \\ \hline
\multicolumn{1}{c|}{} & \multicolumn{9}{c|}{Jaccard similarity} & \multicolumn{9}{c}{PR-AUC} \\ \hline
\begin{tabular}[c]{@{}c@{}}$M_R$\end{tabular} & 2e-4 & 5e-4 & 1e-3 & 2e-3 & 5e-3 & 1e-2 & 2e-2 & 5e-2 & 1e-1 & 2e-4 & 5e-4 & 1e-3 & 2e-3 & 5e-3 & 1e-2 & 2e-2 & 5e-2 & 1e-1 \\ \hline
\ac{er} & \textbf{0.2} & 0.23 & 0.25 & 0.26 & \textbf{0.26} & 0.25 & 0.24 & 0.22 & 0.13 &  \textbf{0.47} & 0.52 & 0.57 & 0.59 & 0.58 & 0.56 & 0.55 & 0.52 & 0.23 \\
\ac{ba} & \textbf{0.24} & 0.3 & \textbf{0.32} & 0.34 & 0.33 & 0.34 & 0.33 & 0.28 & 0.14 & \textbf{0.54} & 0.65 & 0.66 & 0.69 & 0.69 & 0.69 & 0.68 & 0.58 & 0.24 \\
\ac{ws} & \textbf{0.48} & 0.56 & 0.61 & \textbf{0.62} & \textbf{0.6} & 0.59 & 0.57 & 0.56 & 0.45 & 0.79 & \textbf{0.85} & 0.87 & 0.88 & 0.87 & 0.86 & 0.86 & 0.85 & 0.75 \\
\ac{rgg} & 0.63 & \textbf{0.7} & \textbf{0.72} & 0.75 & \textbf{0.73} & 0.75 & 0.74 & 0.71 & 0.27 & 0.84 & 0.88 & 0.88 & 0.89 & \textbf{0.89} & 0.89 & 0.89 & 0.89 & 0.49 \\ \hline
\hline
\end{tabular}
\end{adjustwidth}
\caption{Effect of different mobility rates $M_R$ on the topology inference accuracy of the DTEF model.}
\label{tab: rate_acc}
\end{table*}




\subsubsection*{\label{subsubsec:topinfrealres}Topology inference results on real graphs}
\begin{figure*}[!htbp]
\centering
\includegraphics[width=1\textwidth]{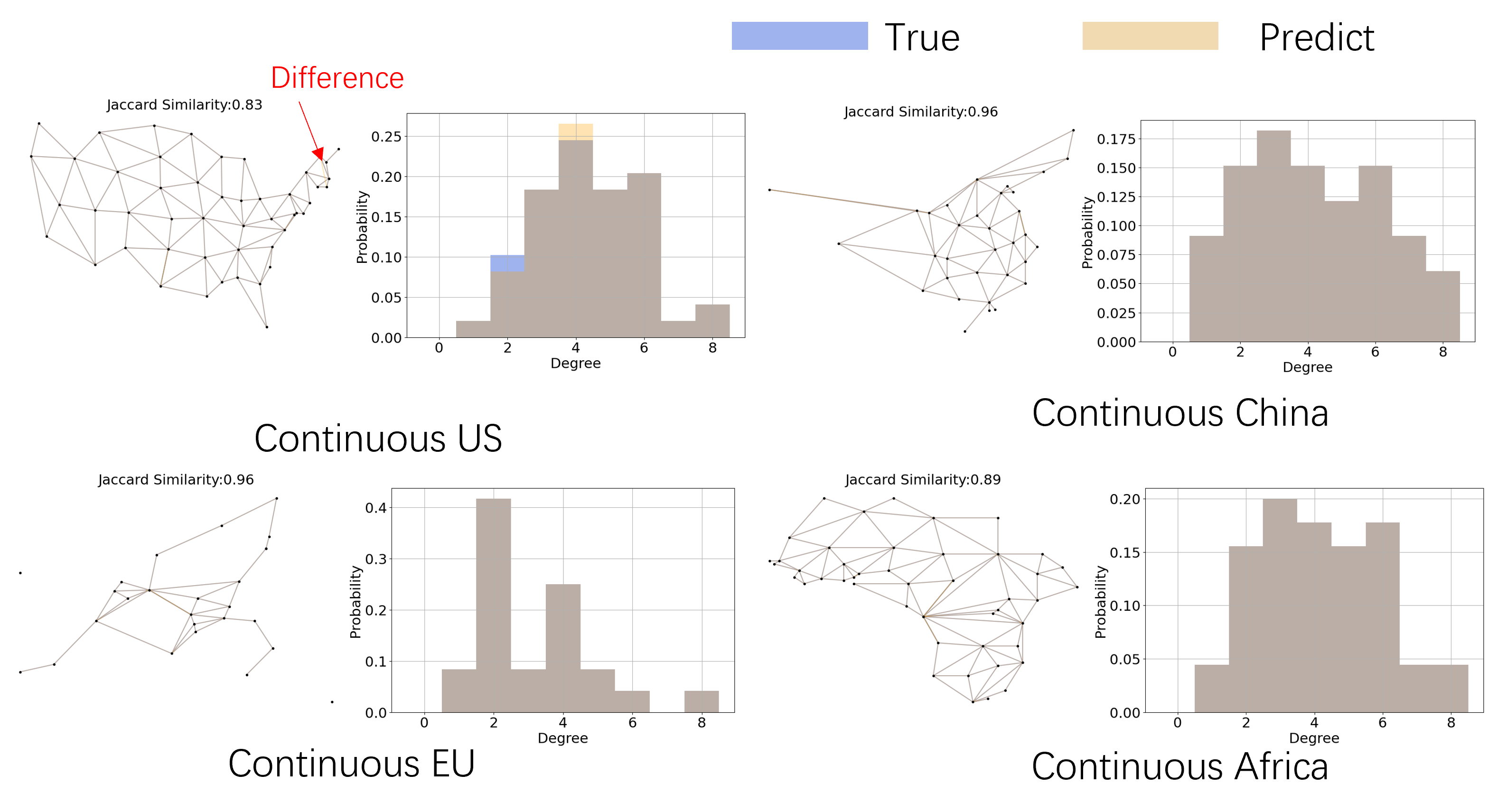}
\caption{The inference accuracy of the DTEF model on real contiguous region graphs: spring layout topology (left) and degree distribution (right). }
\label{fig: real_region}
\end{figure*}

\begin{figure*}[!htbp]
\centering
\includegraphics[width=0.8\textwidth]{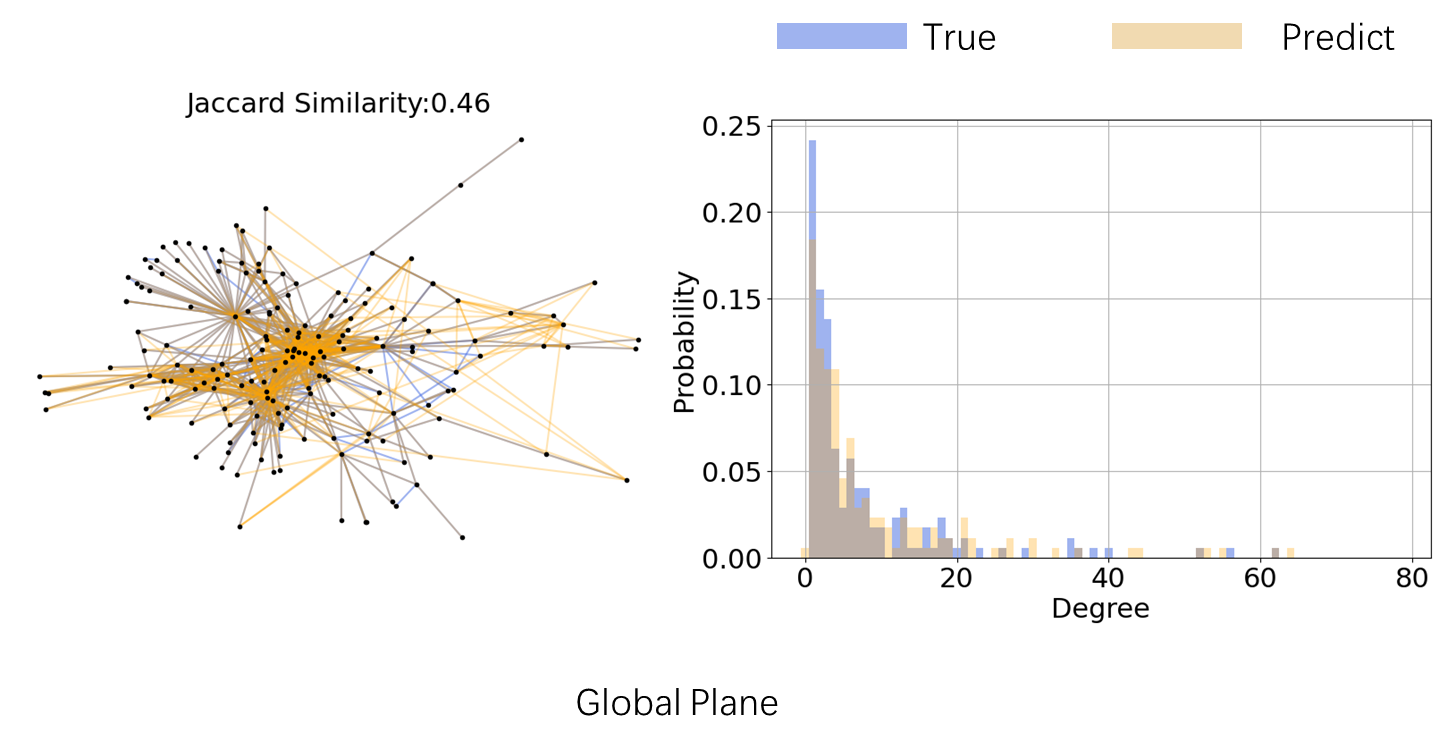}
\caption{The inference accuracy of the DTEF model on real global plane graphs: spring layout topology (left) and degree distribution (right).}
\label{fig: real_global_plane}
\end{figure*}

\begin{figure*}[!htbp]
\centering
\includegraphics[width=1\textwidth]{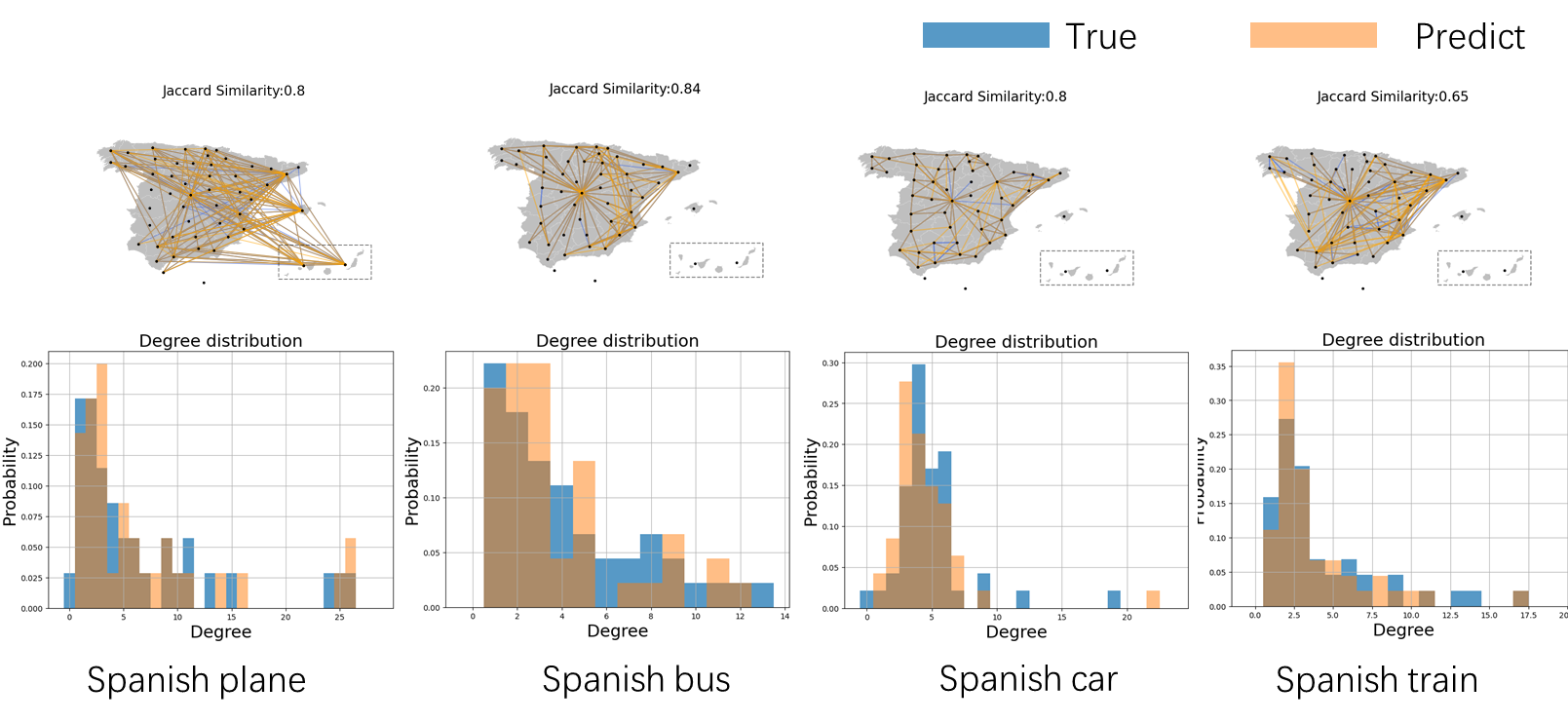}
\caption{The inference accuracy of the DTEF model on real Spanish graphs: spring layout topology (upper Figures) and degree distribution (lower Figures).}
\label{fig: real_spanish}
\end{figure*}

In this appendix we present further experiments of all real graphs and their fitting results (i.e., as in Figure~\ref{fig: pathogen_hst}), as inferred by the DTEF model. Table~\ref{tab: real_results} contains the numerical values of the real graph experimental results, and in Figures~\ref{fig: real_region}-\ref{fig: real_spanish} we present additional experimental results on real graphs.

\begin{table*}[!htbp]
\begin{adjustwidth}{-2.5in}{0in}
\centering
\begin{tabular}{ccccccccccccccccc}
 \hline
 \hline
\multicolumn{1}{l|}{} & \multicolumn{4}{l|}{Spectral   similarity} & \multicolumn{4}{l|}{Pearson correlation} & \multicolumn{4}{l|}{Jaccard similarity}  & \multicolumn{4}{l}{PR-AUC} \\ \hline
$Pathogen$ & \multicolumn{1}{r}{1} & \multicolumn{1}{r}{2} & \multicolumn{1}{r}{3} & \multicolumn{1}{r}{4} & \multicolumn{1}{r}{1} & \multicolumn{1}{r}{2} & \multicolumn{1}{r}{3} & \multicolumn{1}{r}{4} & 1 & 2 & 3 & 4 & \multicolumn{1}{r}{1} & \multicolumn{1}{r}{2} & \multicolumn{1}{r}{3} & \multicolumn{1}{r}{4}  \\ \hline
contiguous\_US & 0.46 & 0.78 & 0.82 & \textbf{0.84} & 0.58 & 0.93 & 0.98 & \textbf{0.98} & 0.26 & 0.69 & \textbf{0.86} & 0.83  & 0.53 & 0.82 & \textbf{0.83} & \textbf{0.83} \\
contiguous\_China & 0.55 & 0.81 & 0.71 & \textbf{0.98} & 0.64 & 0.93 & 0.99 & \textbf{1} & 0.3 & 0.69 & 0.9 & \textbf{0.96}  & 0.57 & 0.82 & \textbf{0.83} & \textbf{0.83} \\
contiguous\_EU & 0.17 & 0.35 & \textbf{0.63} & 0.58 & 0.7 & 0.96 & 0.99 & \textbf{1} & 0.38 & 0.78 & 0.92 & \textbf{0.96}  & 0.66 & 0.8 & \textbf{0.8} & \textbf{0.8} \\
contiguous\_Africa & 0.36 & 0.73 & 0.81 & \textbf{0.85} & 0.53 & 0.9 & 0.96 & \textbf{0.99} & 0.22 & 0.63 & 0.77 & \textbf{0.89}  & 0.44 & 0.81 & \textbf{0.83} & \textbf{0.83} \\
mobility\_german & \textbf{0.21} & 0.13 & 0.08 & 0.04 & 0.05 & 0.32 & 0.4 & \textbf{0.54} & 0 & 0.12 & 0.17 & \textbf{0.22}  & 0.31 & 0.44 & 0.51 & \textbf{0.52} \\
mobility\_US & 0.26 & 0.08 & 0.56 & \textbf{0.77} & 0.33 & 0.93 & 0.95 & \textbf{0.96} & 0.18 & 0.36 & 0.41 & \textbf{0.48}  & 0.88 & 0.92 & 0.93 & \textbf{0.95} \\
Air\_global & 0.02 & 0.11 & 0.06 & \textbf{0.06} & 0.45 & 0.74 & 0.79 & \textbf{0.82} & 0.18 & 0.35 & 0.4 & \textbf{0.46}  & 0.43 & 0.69 & 0.76 & \textbf{0.79} \\
US\_plane\_top\_100 & 0.68 & \textbf{0.72} & 0.5 & 0.35 & 0.54 & 0.69 & 0.74 & \textbf{0.81} & 0.28 & 0.35 & 0.4 & \textbf{0.49} & \multicolumn{1}{r}{0.6} & \multicolumn{1}{r}{0.69} & \multicolumn{1}{r}{0.75} & \multicolumn{1}{r}{\textbf{0.82}} \\
US\_plane\_top\_200 & 0.68 & 0.73 & \textit{\textbf{0.76}} & 0.68 & 0.28 & 0.69 & 0.73 & \textbf{0.78} & 0.17 & 0.33 & 0.37 & \textbf{0.42}  & \multicolumn{1}{r}{0.37} & \multicolumn{1}{r}{0.69} & \multicolumn{1}{r}{0.73} & \multicolumn{1}{r}{\textbf{0.79}} \\
US\_plane\_top\_300 & 0.24 & 0.59 & 0.69 & \textbf{0.7} & 0.08 & 0.69 & 0.72 & \textbf{0.75} & 0.03 & 0.32 & 0.34 & \textbf{0.38}  & \multicolumn{1}{r}{0.09} & \multicolumn{1}{r}{0.67} & \multicolumn{1}{r}{0.71} & \multicolumn{1}{r}{\textbf{0.76}} \\
US\_plane\_top\_400 & 0.35 & \textbf{0.7} & 0.69 & 0.53 & 0.3 & 0.65 & 0.69 & \textbf{0.71} & 0.01 & 0.3 & 0.33 & \textbf{0.35}  & \multicolumn{1}{r}{0.21} & \multicolumn{1}{r}{0.62} & \multicolumn{1}{r}{0.67} & \multicolumn{1}{r}{\textbf{0.68}} \\
Spanish\_bus & \textbf{0.76} & 0.71 & 0.57 & 0.64 & 0.75 & 0.84 & 0.9 & \textbf{0.92} & 0.41 & 0.52 & 0.62 & \textbf{0.69}  & \multicolumn{1}{r}{0.69} & \multicolumn{1}{r}{0.77} & \multicolumn{1}{r}{0.81} & \multicolumn{1}{r}{\textbf{0.82}} \\
Spanish\_car & 0.26 & \textbf{0.5} & 0.34 & 0.44 & 0.51 & 0.71 & 0.85 & \textbf{0.91} & 0.22 & 0.36 & 0.54 & \textbf{0.64} & \multicolumn{1}{r}{0.46} & \multicolumn{1}{r}{0.63} & \multicolumn{1}{r}{0.78} & \multicolumn{1}{r}{\textbf{0.81}} \\
Spanish\_plane & \textbf{0.94} & 0.75 & 0.75 & 0.72 & 0.73 & 0.86 & 0.89 & \textbf{0.92} & 0.46 & 0.62 & 0.67 & \textbf{0.73}  & \multicolumn{1}{r}{0.75} & \multicolumn{1}{r}{0.8} & \multicolumn{1}{r}{0.83} & \multicolumn{1}{r}{\textbf{0.85}} \\
Spanish\_train & 0.56 & 0.55 & 0.77 & \textbf{0.83} & 0.7 & 0.71 & 0.8 & \textbf{0.84} & 0.35 & 0.35 & 0.46 & \textbf{0.52}  & \multicolumn{1}{r}{0.67} & \multicolumn{1}{r}{0.67} & \multicolumn{1}{r}{0.73} & \multicolumn{1}{r}{\textbf{0.76}} \\ \hline
\hline
\end{tabular}
\end{adjustwidth}
\caption{Real-world graph evaluation results for pathogens one to four, as inferred by the DTEF model.}
\label{tab: real_results}
\end{table*}

%
%

\bibliography{main}

\begin{acronym}[ICANN]
    \acro{dl}[DL]{deep learning}
    \acro{dti}[DTI]{deep learning topology inference}
    \acro{efb}[EFB]{epidemic fast-forward-backward}
    \acro{seir}[SEIR]{susceptible-exposed-infectious-recovered}
    \acro{sir}[SIR]{susceptible-infectious-recovered}
    \acro{sird}[SIRD]{susceptible-infectious-recovered-deceased}
    \acro{esc}[ESC]{epidemic sequential computation}
    \acro{fti}[FTI]{fast topology inference}
    \acro{dtef}[DTEF]{deep learning topology inference and epidemic fast-forward-backward}
    \acro{ftef}[FTEF]{fast topology inference and epidemic fast-forward-backward}
    \acro{er}[ER]{Erdos-Renyi random graph}
    \acro{ba}[BA]{Barabasi-Albert scale-free random graph}
    \acro{ws}[WS]{Watts-Strogatz small-world random graph}
    \acro{rgg}[RGG]{random geometric graph}
    \acro{rmse}[RMSE]{root mean square error}
\end{acronym}

\end{document}